\documentclass[letterpaper]{article} %
\usepackage[preprint]{aaai2027}  %
\usepackage[hyphens]{url}  %
\usepackage{graphicx} %
\usepackage{natbib}  %
\usepackage{caption} %
\usepackage{booktabs}
\usepackage{amsfonts}
\usepackage{amsmath}
\usepackage{amssymb}
\usepackage{array}
\usepackage{multirow}
\usepackage{enumitem}
\usepackage{pifont}
\usepackage{algorithm}
\usepackage{algpseudocode}

\title{Beyond Magnitude and Shape:\\ A Direction-Aware Loss for Time Series Forecasting}
\author{
    Seunghan Lee, Jaehoon Lee, Jun Seo, Junhyeok Kang, Sangjun Han, Sungdong Yoo,\\
    Minjae Kim, Tae Yoon Lim, Dongwan Kang, Hwanil Choi, Soonyoung Lee, Wonbin Ahn
}
\affiliations{
    LG AI Research
}

\begin{document}
\maketitle

\begin{abstract}
The \textit{direction} of change---whether a series will move up or down---is often
as important as its exact value in decision-driven applications such as risk
management and financial forecasting. However, most forecasting losses optimize either
point \textit{magnitude} or \textit{shape} and \textit{frequency}
structure, and none explicitly targets the
direction of change. In this paper, we find that MSE-trained forecasters
\textit{fail on the direction of small moves}.
To address this, we propose \textbf{CosDir}, a simple yet effective
\textit{direction-aware} loss that aligns the difference vectors of the prediction
and the target via cosine similarity. Being \textit{scale-invariant}, CosDir keeps
a directional gradient on small moves, re-injecting learning signal
exactly where MSE neglects it. CosDir is a lightweight, plug-in term that attaches to
any backbone without architectural modification. Since the best ratio for mixing the directional
and magnitude terms differs across datasets, we further propose \textbf{CosDir-UW}, an extension
that makes this ratio \textit{adaptive} by learning it during training, matching a
per-dataset tuned weight with no hyperparameter.
We conduct over 100K experiments,
demonstrating that our method
consistently and significantly improves directional accuracy while preserving magnitude
accuracy,
and that it outperforms various loss functions.
Code is available at: \url{https://github.com/seunghan96/cosdir}.
\end{abstract}

\section{Introduction}

Time series (TS) forecasting is widely used across domains such as energy~\citep{kong2019}
and finance~\citep{jiang2021}. A wide range of forecasting models has been developed based on
different architectures, including Transformers~\citep{nie2023,liu2024} and multi-layer
perceptrons (MLPs)~\citep{zeng2023,li2023}. Regardless of the backbone, these models are almost
universally trained by minimizing a \textit{point-wise magnitude} loss such as the mean squared
error (MSE) or mean absolute error (MAE).

\begin{figure*}[t]
\centering
\includegraphics[width=\textwidth]{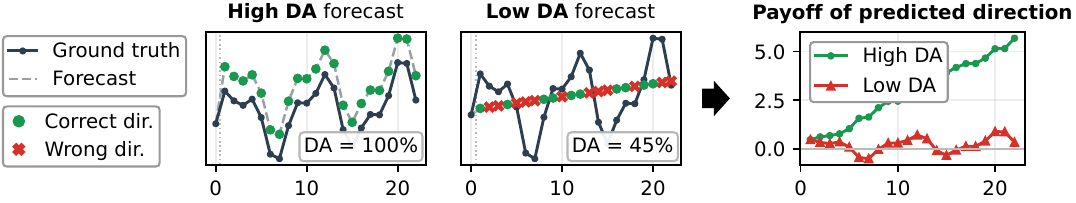}
\caption{
Same MSE, different decision values.
Two forecasts share the \textit{same}
MSE, but the left one (high DA) predicts every change's direction
while the middle one (low DA) does not.
The right panel plots the \textit{payoff} of the predicted direction, and only the high-DA forecast
accumulates it, so it is directional accuracy, not magnitude, that drives the decision.}
\label{fig:decision}
\end{figure*}

A line of work has argued that magnitude losses are insufficient and has proposed
losses that capture the \textit{shape} or \textit{frequency} structure of a forecast.
As summarized in Table~\ref{tab:whitespace}, however,
\textit{none of these objectives explicitly optimizes the direction of change}---i.e., whether the
series moves up or down at each step---even though in many decision-driven applications the
direction ultimately drives the downstream action. As Figure~\ref{fig:decision} shows, two
forecasts can have the \textit{same} MSE yet very different directional accuracy (DA), and only the one
that gets the direction right is useful for decisions.

\begin{table}[t]
\centering
\small
\setlength{\tabcolsep}{4pt}
\begin{tabular}{@{}l|ll@{}}
\toprule
Axis & \textit{Does the forecast\ldots} & Example losses \\
\midrule
Magnitude          & Hit the right value?         & MSE, MAE \\
Shape              & Follow the right shape?      & Soft-DTW, TILDE-Q \\
Frequency          & Match the right frequencies? & FreDF, DBLoss \\
\textbf{Direction} & Move in the right direction? & \textbf{CosDir (Ours)} \\
\bottomrule
\end{tabular}
\caption{Comparison of forecasting losses.
}
\label{tab:whitespace}
\end{table}

\begin{table*}[t]
\centering
\small
\setlength{\tabcolsep}{6pt}
\begin{tabular}{llllc}
\toprule
Method & Training objective & Extra term captures & Optimized axis & Scale-inv. \\
\midrule
MSE & $\lVert\hat{\mathbf{y}}-\mathbf{y}\rVert_2^2$ & --- (Base only) & Magnitude & --- \\
MAE & $\lVert\hat{\mathbf{y}}-\mathbf{y}\rVert_1$ & --- (Base only) & Magnitude & --- \\
DILATE / Soft-DTW & Shape alignment via (soft-)DTW & Temporal shape & Shape & \ding{55} \\
TILDE-Q & $\lVert\sigma(\hat{\mathbf{y}})-\sigma(\mathbf{y})\rVert_1$ & Amplitude / phase invariance & Shape & \ding{55} \\
FreDF & $\lVert\mathcal{F}(\hat{\mathbf{y}})-\mathcal{F}(\mathbf{y})\rVert_1$ & Frequency spectrum & Frequency & \ding{55} \\
DBLoss & $\lVert T(\hat{\mathbf{y}})-T(\mathbf{y})\rVert_1+\lVert S(\hat{\mathbf{y}})-S(\mathbf{y})\rVert_2^2$ & Trend / seasonal decomposition & Frequency & \ding{55} \\
\textbf{CosDir (Ours)} & $1-\cos(\Delta\hat{\mathbf{y}},\Delta\mathbf{y})$ & Direction of change & \textbf{Direction} & \ding{51} \\
\bottomrule
\end{tabular}
\caption{
Various loss functions.
Each augments MSE with a structural term,
and only
CosDir is directional and scale-invariant.}
\label{tab:losses}
\end{table*}

To this end, we propose \textbf{CosDir}, a simple direction-aware loss that aligns the horizon
\textit{difference vectors} of the prediction and the target through cosine similarity. Being
\textit{scale-invariant}, CosDir keeps
a directional gradient on small moves, re-injecting
learning signal exactly where MSE is blind, and it is a lightweight plug-in that attaches to any
backbone and base loss.
Because the best weight for mixing the directional and magnitude terms varies widely across datasets, we further
propose \textbf{CosDir-UW}, an extension that \textit{learns} this ratio via uncertainty weighting.
Across over 100K experiments
spanning 13 datasets, 15 backbones, and up to six horizons, our methods consistently improve DA while
preserving magnitude and outperform a diverse set of
loss functions on direction.
Our main contributions are:
\begin{itemize}[leftmargin=1.2em,itemsep=1pt,topsep=2pt]
\item We identify an overlooked failure mode of magnitude losses: MSE-trained forecasters
\textit{systematically mispredict the direction of small moves}, because their gradient is
dominated by large-magnitude points.
\item We propose \textbf{CosDir}, a simple yet effective direction-aware loss that aligns
prediction and target difference vectors via cosine similarity. Being \textit{scale-invariant},
CosDir keeps a directional
gradient on small moves, where MSE's vanishes. It is a plug-in for
any backbone and base loss.
\item We further propose \textbf{CosDir-UW}, an extension that makes the direction--magnitude
mixing ratio \textit{adaptive} by learning it via uncertainty weighting, motivated by the
observation that the best fixed ratio differs across datasets.
\item We run over 100K experiments across 13 datasets, 15 backbones, and up to six forecasting horizons, showing that CosDir consistently \textit{improves DA without degrading MSE} and outperforms various loss functions.
\end{itemize}
\section{Related Work}

\textbf{TS forecasting models.}
Deep TS forecasting models are commonly grouped into
MLP-based~\citep{zeng2023,chen2023,wang2024timemixer,das2023tide,zhang2022lightts,yi2023frets,zhou2022film},
CNN-based~\citep{wu2023}, and
Transformer-based~\citep{vaswani2017,zhou2021,wu2021,zhou2022,liu2022pyraformer,kitaev2020,liu2022nonstationary,zhang2023crossformer,nie2023,liu2024}
families, along with recurrent designs~\citep{lin2023segrnn}.
Details of all 15 backbones used in this work are deferred to
Appendix~\ref{app:backbones}.

\textbf{Magnitude losses.}
Most TS forecasting models are trained with point-wise magnitude losses such as MSE and MAE,
which measure the average deviation between predicted and true values~\citep{nie2023,liu2024,zeng2023}.
These losses are simple and differentiable, but they are dominated by large-magnitude errors
and are largely insensitive to the direction of low-amplitude changes.

\textbf{Shape and frequency losses.}
A growing body of work argues that magnitude losses fail to capture temporal structure and
proposes structure-aware alternatives, which we summarize in Table~\ref{tab:losses}.
Soft-DTW~\citep{cuturi2017} and DILATE~\citep{leguen2019}
align the temporal \textit{shape} of forecasts via differentiable dynamic time warping.
TILDE-Q~\citep{lee2024tildeq} enforces invariance to amplitude, phase, and uniform shifts.
FreDF~\citep{wang2024fredf} matches the \textit{frequency} spectrum of prediction and target,
and DBLoss~\citep{qiu2025dbloss} decomposes the series into trend and seasonal components and
computes separate losses on each. First-order difference terms also appear inside several shape
losses to encourage local smoothness, but they are \textit{magnitude-weighted}: their contribution
vanishes for small moves, leaving directional errors uncorrected.

\textbf{Directional metrics, losses, and models.}
Directional (or ``hit-rate'') metrics are used to \textit{evaluate} forecasts~\citep{lee2024tildeq},
but rarely serve as a \textit{training} objective because the sign function is
non-differentiable. The Mean Absolute Directional Loss (MADL)~\citep{michankow2024madl} is a
finance-specific exception that weights
the realized return by the sign agreement between
prediction and target.
Yet its discrete sign operator \textit{limits its effectiveness as a differentiable objective}.
Rank- and correlation-based surrogates, made trainable by differentiable sorting and
ranking~\citep{blondel2020fast}, likewise align change trajectories monotonically rather than pointwise.
CosDir instead casts direction as a \textit{smooth} cosine alignment that is \textit{scale-invariant}
and
differentiable, supplying a directional gradient at every
magnitude.
Beyond loss design, CaReTS~\citep{yao2025carets} is a multi-task decomposition that
disentangles direction and magnitude and balances them
at the
\emph{architecture} level with dual streams. CosDir instead injects direction through the loss
and remains a single-stream, backbone-agnostic plug-in. This aligns with decision-focused learning,
where surrogates that preserve decision-relevant information outperform pure prediction-error
losses~\citep{elmachtoub2022spo}.
\section{Preliminaries}

\begin{figure*}[!t]
\centering
\includegraphics[width=\textwidth]{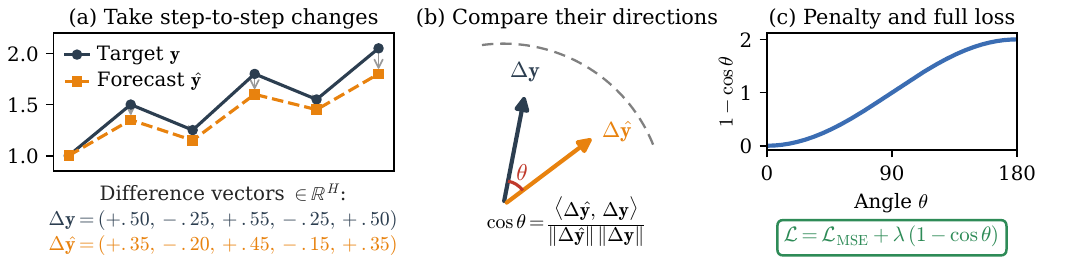}
\caption{Overview of CosDir. (a) For each channel we take the step-to-step changes of the forecast
and the target, which stack into the horizon-length difference vectors $\Delta\hat{\mathbf{y}}$ and
$\Delta\mathbf{y}$ (written out for a five-step example). (b) We compare their directions through
cosine similarity, which yields the angle $\theta$ between the two change vectors and depends only
on direction, not magnitude. (c) The penalty $1-\cos\theta$ is added to the MSE, so it is $0$ when
the change directions align and grows toward $2$ as they diverge.}
\label{fig:method}
\end{figure*}

\textbf{TS forecasting.}
In TS forecasting, a model predicts the future values $\mathbf{y}=(\mathbf{x}_{L+1},\ldots,\mathbf{x}_{L+H})$
given a lookback window $\mathbf{x}=(\mathbf{x}_1,\ldots,\mathbf{x}_L)$. Each $\mathbf{x}_i\in\mathbb{R}^C$
represents values at time step $i$, where $L$, $H$, and $C$ denote the size of the lookback window,
the forecast horizon, and the number of channels. A model $f_\theta$ produces a
prediction $\hat{\mathbf{y}}=f_\theta(\mathbf{x})\in\mathbb{R}^{H\times C}$.

\textbf{Direction of change.}
For channel $c$, we define the \textit{one-step change} at horizon step $h$ as the first difference
$\Delta y_{h,c}=y_{h,c}-y_{h-1,c}$, with $y_{0,c}$ set to the last observed value $x_{L,c}$ so that
the direction of the first predicted step is included. The \textit{direction} at step $h$ is then
$\operatorname{sign}(\Delta y_{h,c})$, and analogously $\operatorname{sign}(\Delta\hat{y}_{h,c})$ for
the prediction.

\textbf{Evaluation metrics.}
We assess forecasts along two complementary axes. \textit{Magnitude} is measured by the standard
MSE and MAE, which quantify how close the predicted
values are to the targets. \textit{Direction} is measured by DA, the
fraction of horizon steps at which the predicted and true change directions agree,
\begin{equation}
\text{DA}=\frac{1}{HC}\sum_{h=1}^{H}\sum_{c=1}^{C}
\mathbf{1}\!\left[\operatorname{sign}(\Delta\hat{y}_{h,c})=\operatorname{sign}(\Delta y_{h,c})\right],
\label{eq:da}
\end{equation}
where $\mathbf{1}[\cdot]$ is the indicator function. Higher DA is better, with $0.5$ corresponding to
chance for a balanced up/down split. DA is our primary directional metric and is orthogonal to
magnitude error, as a forecast can be accurate in value yet wrong in direction, and vice versa.
\section{Method}

We introduce \textbf{CosDir}, a simple yet effective direction-aware loss, illustrated in
Figure~\ref{fig:method}. We first analyze \textit{why} magnitude losses fail on direction
(Sec.~``Why MSE Neglects Direction''), then present CosDir and its scale-invariance property.

\subsection{Why MSE Neglects Direction}
The gradient of the MSE with respect to a prediction is proportional to the signed error,
$\partial \mathcal{L}_{\text{MSE}}/\partial \hat{y}_{h,c} \propto (\hat{y}_{h,c}-y_{h,c})$.
As a result, optimization is dominated by \textit{large-magnitude} points, and the contribution
of a step to the total gradient scales with the size of its error rather than with whether its
\textit{sign} is correct. Low-amplitude changes therefore receive negligible gradient, so the
model has little incentive to get their direction right. Empirically,
this produces a DA that increases monotonically with the true move size and
collapses toward (or below) chance for the smallest moves, where most directional errors occur.
This motivates a \textit{scale-invariant} directional objective
whose directional gradient does not vanish on low-amplitude moves, unlike the squared error.

\subsection{CosDir: Scale-Invariant Directional Alignment}
Let $\Delta\hat{\mathbf{y}}_{c}=(\Delta\hat{y}_{1,c},\ldots,\Delta\hat{y}_{H,c})\in\mathbb{R}^{H}$
and $\Delta\mathbf{y}_{c}\in\mathbb{R}^{H}$ denote the horizon \textit{difference vectors} of the
prediction and the target for channel $c$, respectively.
CosDir aligns their \textit{orientation} through the cosine similarity:
\begin{equation}
\mathcal{L}_{\text{CosDir}}
= \mathcal{L}_{\text{MSE}}
+ \lambda \cdot \frac{1}{C}\sum_{c=1}^{C}
\left(1 - \frac{\langle \Delta\hat{\mathbf{y}}_{c},\, \Delta\mathbf{y}_{c}\rangle}
{\lVert \Delta\hat{\mathbf{y}}_{c}\rVert\,\lVert \Delta\mathbf{y}_{c}\rVert + \epsilon}\right),
\label{eq:cosdir}
\end{equation}
where $\langle\cdot,\cdot\rangle$ is the inner product over the horizon,
$\epsilon$ is a small
constant,
and $\lambda$ is the
hyperparameter
balancing the
directional term against the base loss. The cosine term is minimized ($=0$) when the
two
change trajectories point in the same direction and maximized ($=2$) when they are opposed.
CosDir has the three properties below.

\textbf{Property 1: Scale invariance.}
The key property of Eq.~\eqref{eq:cosdir} is that the cosine term depends only on the
\textit{orientation} of the difference vectors, not their magnitude:
$\cos(a\,\Delta\hat{\mathbf{y}}_{c}, b\,\Delta\mathbf{y}_{c})=\cos(\Delta\hat{\mathbf{y}}_{c},\Delta\mathbf{y}_{c})$
for any $a,b>0$.
Consequently, the objective is invariant to the overall amplitude of a window or channel, removing the magnitude bias that leads MSE to ignore low-amplitude series.
Within a horizon, the cosine still weights each step by its size, but only linearly and after normalization, so unlike the squared error it keeps a directional gradient on small moves.
This is in contrast to first-difference
terms of the form
$\lVert\Delta\hat{\mathbf{y}}-\Delta\mathbf{y}\rVert^2$, whose gradient is \textit{magnitude-weighted}
and vanishes for small moves.
Appendix~\ref{app:grad} derives the gradient of the cosine term and shows how its directional signal distributes across horizon steps.

\textbf{Property 2: Differentiability.}
Because the sign function is \textit{non-differentiable}, directional objectives are often cast as
hard classification and optimized with surrogate gradients. CosDir instead replaces the sign with a
\textit{smooth} cosine surrogate: the first-difference operator is linear in the prediction, the
cosine term is differentiable everywhere except where a predicted difference vector is exactly
zero, and the stabilizer $\epsilon$ keeps the loss and its gradient bounded even at that point.
CosDir is thus trained end-to-end without a straight-through estimator.
When a target window is nearly flat ($\lVert\Delta\mathbf{y}\rVert\!\approx\!0$), its direction is
undefined, but such windows carry little directional information, so their gradient contribution is
naturally small. Masking or confidence-weighting them by $\lVert\Delta\mathbf{y}\rVert$ is a
compatible refinement we leave to future work.

\textbf{Property 3: Applicability.}
CosDir is broadly applicable in three ways. \textit{(3-1) Complementary}: it constrains only the
orientation of the change trajectory without competing with the magnitude objective, so that MSE is
preserved. \textit{(3-2) Plug-in}: it is added to any base loss and attaches to any backbone.
\textit{(3-3) Lightweight}: it adds only $O(HC)$ computation,
i.e., $O(1)$ overhead relative to a forward pass.

\subsection{CosDir-UW: Balancing Magnitude and Direction}

\textbf{Motivation.}
In CosDir, the weight $\lambda$ sets \textit{how strongly} direction is emphasized relative to
magnitude, and it is \textit{fixed} across datasets. However, the most effective ratio can depend
on the data. Empirically, the best
fixed $\lambda$ varies by an order of magnitude across datasets (from $0.3$ to $3.0$),
so no single global ratio is optimal, as further discussed in Appendix~\ref{app:uw}.

\begin{figure}[t]
\centering
\includegraphics[width=\columnwidth]{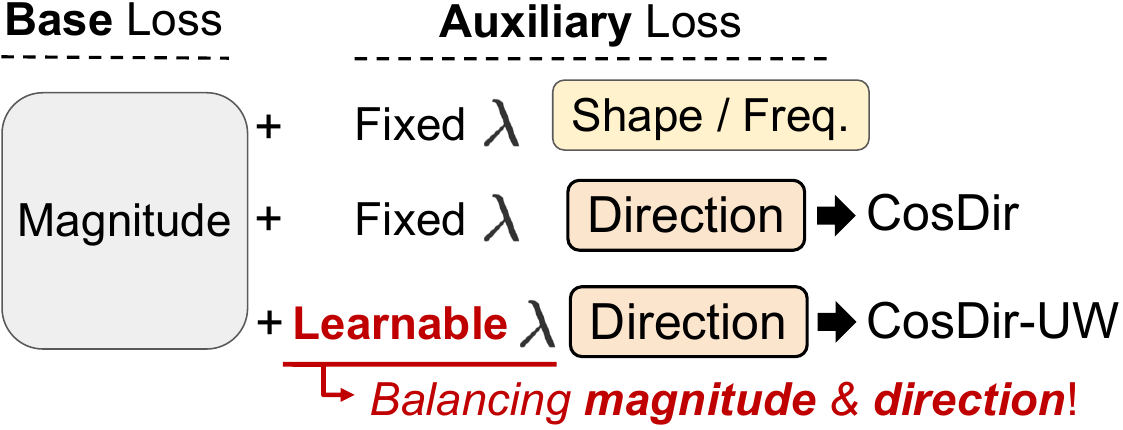}
\caption{
CosDir vs.\ CosDir-UW. CosDir mixes direction and magnitude with a fixed weight $\lambda$,
whereas CosDir-UW makes $\lambda$ learnable so the balance adapts by dataset.
}
\label{fig:cosuw}
\end{figure}

\textbf{Formulation.}
To let the model \textit{learn} this ratio from data rather than fix it as a hyperparameter,
we propose \textbf{CosDir-UW} (Figure~\ref{fig:cosuw}), which casts the magnitude and direction
terms as two objectives with unknown, learnable
confidence, following
homoscedastic \textbf{U}ncertainty \textbf{W}eighting~\citep{kendall2018}.
Introducing two scalar log-variance parameters $s_1,s_2\in\mathbb{R}$, the objective is
\begin{equation}
\mathcal{L}_{\text{UW}}
= e^{-s_1}\,\mathcal{L}_{\text{MSE}}
+ e^{-s_2}\,\mathcal{L}_{\text{dir}}
+ \tfrac{1}{2}\left(s_1+s_2\right),
\label{eq:uw}
\end{equation}
where $\mathcal{L}_{\text{dir}}=\frac{1}{C}\sum_{c=1}^{C}\big(1-\cos(\Delta\hat{\mathbf{y}}_{c},\Delta\mathbf{y}_{c})\big)$
is the \textit{same} cosine penalty as in Eq.~\eqref{eq:cosdir}. The two precisions $e^{-s_1}$ and
$e^{-s_2}$ are learned jointly with the forecasting model.

\textbf{How the balance is learned.}
Because each precision $e^{-s_i}$ can shrink its loss toward zero, the last term is essential:
without it the trivial optimum drives $s_1,s_2\!\to\!\infty$ and both losses vanish. The regularizer
$\tfrac{1}{2}(s_1+s_2)$ penalizes
this,
and setting $\partial\mathcal{L}_{\text{UW}}/\partial s_i=0$
gives a finite optimum at which each precision equals the inverse of its own loss, so a term that is
easier to reduce is automatically up-weighted. The effective directional weight is
$\lambda_{\text{eff}}=e^{\,s_1-s_2}$, a ratio the optimizer discovers per dataset
instead of being manually specified.

\textbf{Properties.}
CosDir-UW inherits the scale invariance and almost-everywhere differentiability of CosDir, as
$\mathcal{L}_{\text{dir}}$ is left unchanged. It adds only \textit{two} scalar parameters and the
same $O(HC)$ direction computation, remains a backbone-agnostic plug-in, and---crucially---exposes
\textit{no} loss hyperparameter: the identical objective is applied unchanged on every dataset.
\section{Experiments}

\subsection{Experimental Settings}

\begin{table}[t]
\centering
\small
\setlength{\tabcolsep}{4pt}
\begin{tabular}{lcc|lcc}
\toprule
\multicolumn{3}{c|}{\textit{Financial}} & \multicolumn{3}{c}{\textit{General}} \\
\cmidrule(lr){1-3}\cmidrule(l){4-6}
Dataset & $C$ & Frequency & Dataset & $C$ & Frequency \\
\midrule
RealVar   & 27 & Daily & ETTh1       & 7   & Hourly \\
RealVol   & 27 & Daily & ETTh2       & 7   & Hourly \\
RealVol60 & 27 & Daily & ETTm1       & 7   & 15-Min \\
AbsRet    & 27 & Daily & ETTm2       & 7   & 15-Min \\
StockVol  & 24 & Daily & Weather     & 21  & 10-Min \\
          &    &       & Solar       & 137 & Hourly \\
          &    &       & Traffic     & 862 & Hourly \\
          &    &       & Electricity & 321 & Hourly \\
\bottomrule
\end{tabular}
\caption{Statistics of financial \& general-domain datasets. In the wide result tables the
financial datasets are abbreviated RVar, RVol, RVol60, AbsR, and SVol, and Electricity as Elec.}
\label{tab:datasets}
\end{table}

\newcommand{\bbcell}[2]{\shortstack[l]{#1\\ \citep{#2}}}
\begin{table}[t]
\centering
\small
\setlength{\tabcolsep}{4pt}
\begin{tabular}{@{}l|l|l@{}}
\toprule
Transformer & MLP & CNN \& RNN \\
\midrule
\bbcell{PatchTST}{nie2023}     & \bbcell{DLinear}{zeng2023}            & \bbcell{TimesNet}{wu2023} \\
\bbcell{iTransformer}{liu2024} & \bbcell{TimeMixer}{wang2024timemixer} & \bbcell{SegRNN}{lin2023segrnn} \\
\bbcell{Autoformer}{wu2021}    & \bbcell{TSMixer}{chen2023}            & \\
\bbcell{FEDformer}{zhou2022}   & \bbcell{LightTS}{zhang2022lightts}    & \\
\bbcell{Pyraformer}{liu2022pyraformer} & \bbcell{FiLM}{zhou2022film}   & \\
\bbcell{Crossformer}{zhang2023crossformer} & \bbcell{TiDE}{das2023tide} & \\
                               & \bbcell{FreTS}{yi2023frets}           & \\
\bottomrule
\end{tabular}
\caption{Forecasting models grouped by architecture family.}
\label{tab:backbones}
\end{table}

\begin{table*}[!t]
\centering
\small
\setlength{\tabcolsep}{1.2pt}
\begin{tabular}{l|cccccc}
\toprule
\cmidrule(lr){1-7}
\textit{Financial}
& 12 & 24 & 36 & 48 & 60 & 72 \\
\cmidrule{1-1}
\cmidrule(lr){2-2}\cmidrule(lr){3-3}\cmidrule(lr){4-4}\cmidrule(lr){5-5}\cmidrule(lr){6-6}\cmidrule(lr){7-7}
RealVar & 53.02/\underline{55.13}$^{*}$/\textbf{55.46}$^{*}$ & 52.19/\underline{53.73}$^{*}$/\textbf{54.20}$^{*}$ & 51.30/\underline{52.33}$^{*}$/\textbf{52.48}$^{*}$ & 50.99/\underline{51.44}$^{*}$/\textbf{51.77}$^{*}$ & 50.79/\underline{51.31}$^{*}$/\textbf{51.41}$^{*}$ & 50.65/\underline{51.01}$^{*}$/\textbf{51.12}$^{*}$ \\
RealVol & 52.78/\underline{54.62}$^{*}$/\textbf{54.94}$^{*}$ & 52.01/\underline{53.63}$^{*}$/\textbf{53.79}$^{*}$ & 51.21/\underline{52.08}$^{*}$/\textbf{52.23}$^{*}$ & 50.94/\underline{51.52}$^{*}$/\textbf{51.64}$^{*}$ & 50.75/\underline{51.20}$^{*}$/\textbf{51.26}$^{*}$ & 50.62/\underline{50.92}$^{*}$/\textbf{51.02}$^{*}$ \\
RealVol60 & 50.91/\underline{52.51}$^{*}$/\textbf{52.70}$^{*}$ & 51.01/\underline{52.67}$^{*}$/\textbf{52.82}$^{*}$ & 50.94/\underline{52.53}$^{*}$/\textbf{52.74}$^{*}$ & 50.89/\underline{52.58}$^{*}$/\textbf{52.69}$^{*}$ & 50.85/\underline{52.25}$^{*}$/\textbf{52.50}$^{*}$ & 50.67/\underline{51.65}$^{*}$/\textbf{51.82}$^{*}$ \\
StockVol & 52.17/\underline{53.77}$^{*}$/\textbf{54.07}$^{*}$ & 51.53/\underline{52.85}$^{*}$/\textbf{53.13}$^{*}$ & 50.93/\underline{51.60}$^{*}$/\textbf{51.82}$^{*}$ & 50.71/\underline{51.06}$^{*}$/\textbf{51.34}$^{*}$ & 50.59/\underline{50.89}$^{*}$/\textbf{50.99}$^{*}$ & 50.47/\underline{50.72}$^{*}$/\textbf{50.79}$^{*}$ \\
AbsRet & 51.86/\underline{52.01}$^{*}$/\textbf{52.07}$^{*}$ & 50.94/\underline{51.03}$^{*}$/\textbf{51.06}$^{*}$ & 50.67/\textbf{50.73}$^{*}$/\textbf{50.73}$^{*}$ & 50.55/\textbf{50.56}/\textbf{50.56} & 50.45/\textbf{50.49}$^{*}$/\underline{50.48} & \textbf{50.42}/\textbf{50.42}/50.41 \\
\midrule
\cmidrule(lr){1-7}
\textit{General}
& 96 & 192 & 336 & 720 & \multicolumn{2}{|c}{} \\
\cmidrule{1-1}
\cmidrule(lr){2-2}\cmidrule(lr){3-3}\cmidrule(lr){4-4}\cmidrule(lr){5-5}
ETTh1 & 58.67/\underline{60.10}$^{*}$/\textbf{60.75}$^{*}$ & 57.66/\underline{59.41}$^{*}$/\textbf{59.68}$^{*}$ & 56.84/\underline{58.50}$^{*}$/\textbf{58.84}$^{*}$ & 55.76/\underline{57.23}$^{*}$/\textbf{57.59}$^{*}$ & \multicolumn{2}{|c}{\multirow{8}{*}[1.1em]{%
\setlength{\tabcolsep}{3pt}\renewcommand{\arraystretch}{1.15}%
\begin{tabular}{@{}lccc@{}}
\multicolumn{4}{c}{\textbf{Summary: Average DA (\%)}}\\
\cmidrule(lr){1-4}
 & Fin. & Gen. & All \\
\cmidrule(lr){2-4}
MSE & 51.13 & 59.80 & 55.60 \\
$+$CosDir & \underline{51.97} & \underline{61.90} & \underline{57.10} \\
$+$CosDir-UW & \textbf{52.13} & \textbf{62.19} & \textbf{57.32} \\
\end{tabular}}} \\
ETTh2 & 54.83/\underline{57.90}$^{*}$/\textbf{58.17}$^{*}$ & 54.00/\underline{56.68}$^{*}$/\textbf{57.02}$^{*}$ & 53.50/\underline{55.98}$^{*}$/\textbf{56.39}$^{*}$ & 52.93/\underline{55.32}$^{*}$/\textbf{55.93}$^{*}$ & \multicolumn{2}{|c}{} \\
ETTm1 & 54.16/\underline{55.17}$^{*}$/\textbf{55.24}$^{*}$ & 54.05/\underline{54.90}$^{*}$/\textbf{54.93}$^{*}$ & 53.77/\underline{54.51}$^{*}$/\textbf{54.54}$^{*}$ & 53.34/\textbf{54.01}$^{*}$/\underline{53.99}$^{*}$ & \multicolumn{2}{|c}{} \\
ETTm2 & 53.02/\underline{54.69}$^{*}$/\textbf{54.74}$^{*}$ & 52.75/\underline{54.16}$^{*}$/\textbf{54.34}$^{*}$ & 52.43/\underline{53.78}$^{*}$/\textbf{53.97}$^{*}$ & 52.26/\underline{53.56}$^{*}$/\textbf{53.77}$^{*}$ & \multicolumn{2}{|c}{} \\
Weather & 52.42/\underline{54.07}$^{*}$/\textbf{54.43}$^{*}$ & 52.47/\underline{54.26}$^{*}$/\textbf{54.53}$^{*}$ & 52.32/\underline{54.07}$^{*}$/\textbf{54.31}$^{*}$ & 52.34/\underline{54.03}$^{*}$/\textbf{54.29}$^{*}$ & \multicolumn{2}{|c}{} \\
Solar & 68.05/\underline{73.14}$^{*}$/\textbf{73.57}$^{*}$ & 67.26/\underline{73.00}$^{*}$/\textbf{73.39}$^{*}$ & 66.63/\underline{72.34}$^{*}$/\textbf{72.50}$^{*}$ & 65.75/\underline{71.47}$^{*}$/\textbf{71.91}$^{*}$ & \multicolumn{2}{|c}{} \\
Traffic & 72.22/\underline{73.81}$^{*}$/\textbf{74.10}$^{*}$ & 72.78/\underline{74.39}$^{*}$/\textbf{74.68}$^{*}$ & 72.67/\underline{74.26}$^{*}$/\textbf{74.72}$^{*}$ & 71.60/\underline{73.36}$^{*}$/\textbf{73.74}$^{*}$ & \multicolumn{2}{|c}{} \\
Electricity & 70.08/\underline{71.28}$^{*}$/\textbf{71.62}$^{*}$ & 69.15/\underline{70.77}$^{*}$/\textbf{71.15}$^{*}$ & 68.99/\underline{70.41}$^{*}$/\textbf{70.74}$^{*}$ & 68.95/\underline{70.27}$^{*}$/\textbf{70.53}$^{*}$ & \multicolumn{2}{|c}{} \\
\bottomrule
\end{tabular}
\caption{Comparison of MSE with MSE\,$+$\,CosDir and MSE\,$+$\,CosDir-UW. We report DA (\%) at each
$H$, averaged over the 15 backbones and 5 seeds, with each cell reading
MSE\,/\,(MSE\,$+$\,CosDir)\,/\,(MSE\,$+$\,CosDir-UW). The best value is in bold and the second
best is \underline{underlined}, and $^{*}$ marks $p<10^{-3}$ under a paired one-sided Wilcoxon test of that method
against MSE.}
\label{tab:main}

\vspace{0.7em}
\setlength{\tabcolsep}{1.5pt}
\begin{tabular}{l|ccccc|cccccccc|c}
\toprule
\multirow{2.5}{*}{vs.\ Prior work} & \multicolumn{5}{c|}{\textit{Financial}} & \multicolumn{8}{c|}{\textit{General}} & \multirow{2.5}{*}{Avg.} \\
\cmidrule(lr){2-6}\cmidrule(lr){7-14}
 & RVar & RVol & RVol60 & SVol & AbsR & ETTh1 & ETTh2 & ETTm1 & ETTm2 & Weather & Solar & Traffic & Elec. & \\
\midrule
MSE & 51.16 & 51.09 & 50.84 & 50.84 & 50.77 & 57.23 & 53.82 & 53.83 & 52.61 & 52.39 & 66.93 & 72.42 & 69.52 & 56.42 \\
+\,FreDF & 51.38 & 51.29 & 51.23 & 51.19 & 50.63 & 57.29 & 55.64 & 53.91 & 53.71 & \underline{54.36} & 71.39 & 72.79 & 69.87 & 57.29 \\
+\,TILDE-Q & 51.18 & 51.13 & 50.79 & 50.84 & 50.61 & 57.23 & 53.82 & 53.83 & 52.62 & 52.38 & 66.97 & 72.46 & 69.52 & 56.42 \\
+\,DBLoss & 50.95 & 50.91 & 50.58 & 50.71 & 50.64 & 56.87 & 54.32 & 53.91 & 53.11 & 52.53 & 65.19 & 71.61 & 68.95 & 56.18 \\
+\,MADL & 51.29 & 51.20 & 50.67 & 50.91 & 50.58 & 57.41 & 53.97 & 53.82 & 52.69 & 51.59 & 66.98 & 71.31 & 69.29 & 56.29 \\
\textbf{+\,CosDir (Ours)} & \underline{51.91} & \underline{51.79} & \underline{52.28} & \underline{51.40} & \underline{50.82} & \underline{58.81} & \underline{56.47} & \underline{54.65} & \underline{54.05} & 54.11 & \underline{72.49} & \underline{73.97} & \underline{70.95} & \underline{57.98} \\
\textbf{+\,CosDir-UW (Ours)} & \textbf{52.56} & \textbf{52.32} & \textbf{52.42} & \textbf{51.89} & \textbf{50.83} & \textbf{59.22} & \textbf{56.88} & \textbf{54.67} & \textbf{54.21} & \textbf{54.39} & \textbf{72.84} & \textbf{74.27} & \textbf{71.27} & \textbf{58.29} \\
\bottomrule
\end{tabular}
\caption{Comparison with various loss functions.}
\label{tab:sota}

\vspace{0.7em}
\begin{tabular}{l|ccccc|cccccccc|c}
\toprule
\multirow{2.5}{*}{vs.\ Our variants} & \multicolumn{5}{c|}{\textit{Financial}} & \multicolumn{8}{c|}{\textit{General}} & \multirow{2.5}{*}{Avg.} \\
\cmidrule(lr){2-6}\cmidrule(lr){7-14}
 & RVar & RVol & RVol60 & SVol & AbsR & ETTh1 & ETTh2 & ETTm1 & ETTm2 & Weather & Solar & Traffic & Elec. & \\
\midrule
MSE & 51.16 & 51.09 & 50.84 & 50.84 & 50.77 & 57.23 & 53.82 & 53.83 & 52.61 & 52.39 & 66.93 & 72.42 & 69.52 & 56.42 \\
+\,First-Difference & 51.51 & 51.43 & 50.85 & 51.06 & 50.59 & 57.66 & 54.45 & 54.28 & 53.02 & 52.76 & 68.60 & 73.63 & 70.16 & 56.92 \\
+\,Magnitude-Weighted Sign & 51.49 & 51.38 & 50.85 & 51.03 & 50.62 & 57.91 & 54.26 & 53.77 & 52.55 & 52.03 & 67.09 & 73.07 & 70.21 & 56.63 \\
+\,Sign Classification (BCE) & 51.90 & 51.77 & 51.15 & \underline{51.43} & 50.62 & 57.94 & 55.72 & \textbf{55.16} & 53.89 & 54.07 & 71.49 & \textbf{75.19} & 70.54 & 57.76 \\
\textbf{+\,CosDir (Ours)} & \underline{51.91} & \underline{51.79} & \underline{52.28} & 51.40 & \underline{50.82} & \underline{58.81} & \underline{56.47} & 54.65 & \underline{54.05} & \underline{54.11} & \underline{72.49} & 73.97 & \underline{70.95} & \underline{57.98} \\
\textbf{+\,CosDir-UW (Ours)} & \textbf{52.56} & \textbf{52.32} & \textbf{52.42} & \textbf{51.89} & \textbf{50.83} & \textbf{59.22} & \textbf{56.88} & \underline{54.67} & \textbf{54.21} & \textbf{54.39} & \textbf{72.84} & \underline{74.27} & \textbf{71.27} & \textbf{58.29} \\
\bottomrule
\end{tabular}
\caption{Ablation with our non-scale-invariant direction-aware variants.}
\label{tab:ablation}
\end{table*}

\textbf{Datasets.}
We evaluate on 13 datasets in two families (Table~\ref{tab:datasets}). Our \textit{primary} domain
is \textit{finance}, represented by five multivariate datasets (RealVar, RealVol, RealVol60,
AbsRet, and StockVol), where the direction of change (whether risk rises or falls) is the
decision-relevant quantity.
For generality we also use eight standard benchmarks: ETTh1, ETTh2, ETTm1, ETTm2~\citep{zhou2021},
Weather, Solar, Traffic, and Electricity~\citep{wu2021}.
Details
are
in Appendix~\ref{app:data}.

\textbf{Backbones.}
To confirm that CosDir is backbone-agnostic, we evaluate on 15 forecasting models spanning
Transformer-, MLP-, CNN-, and RNN-based families, listed in Table~\ref{tab:backbones}.

\textbf{Protocol.}
We use a lookback $L=96$, with horizons $H\in\{96,192,336,720\}$ on the general benchmarks and
$H\in\{12,24,36,48,60,72\}$ on the financial datasets, where volatility direction is most
predictable. MSE is the base loss and CosDir an \textit{auxiliary} term on top, so
``w/o'' and ``w/ CosDir'' denote MSE and MSE\,$+$\,CosDir. Unless noted, $\lambda=0.5$. All numbers
are the mean over \textit{5 random seeds} (per-seed standard deviations in Appendix~\ref{app:std}).
Our primary metric is DA, with magnitude preservation verified in Figure~\ref{fig:mse}.
Significance is assessed with a paired one-sided Wilcoxon test on matched (dataset, backbone,
horizon, seed) cells, with the threshold and a runtime comparison in Appendix~\ref{app:protocol}.

\subsection{Main Results}

We assess
CosDir and CosDir-UW
along three axes---(1) \textbf{effectiveness}, (2) \textbf{competitiveness}, and (3) \textbf{the source of
their gains}---
spanning
\textit{15 backbones} and \textit{13 datasets}:
\begin{itemize}[leftmargin=1.6em,itemsep=1pt,topsep=2pt]
\item (1) MSE vs.\ MSE\,$+$\,CosDir/CosDir-UW
\item (2) Comparison with various loss functions
\item (3) Ablation of the direction-aware term
\end{itemize}

\textbf{MSE vs.\ MSE\,$+$\,CosDir/CosDir-UW.}
To demonstrate that
our methods
improve directional accuracy
across backbones,
datasets, and horizons, we compare the
MSE baseline against MSE augmented with each.
As shown in Table~\ref{tab:main}, CosDir raises DA on both the financial datasets and the general
benchmarks, with the clearest gains at the short horizons where the direction of risk is most
predictable. CosDir-UW brings an additional gain over CosDir, demonstrating the effectiveness of a
learnable $\lambda$.
As shown in Figure~\ref{fig:mse},
these directional gains \textit{do not trade off magnitude accuracy (MSE)}. Note that the return-based series are near chance:
directional hit rates on raw stock prices are known to be
hard to push much above $50\%$,
consistent with the difficulty of stock-price forecasting under market efficiency~\citep{fama1970efficient}.

\begin{figure*}[t]
\centering
\includegraphics[width=0.45\textwidth]{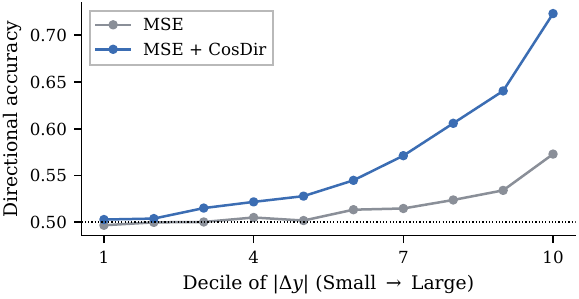}\hspace{2.5em}%
\includegraphics[width=0.45\textwidth]{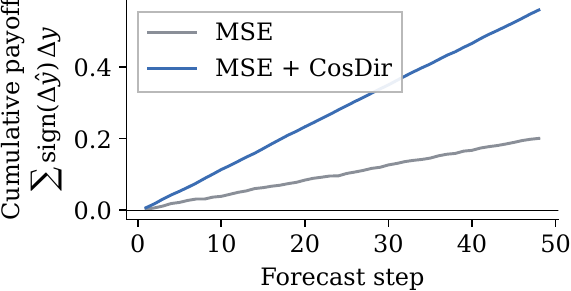}
\caption{Directional gain grows with move size and turns into payoff. \textit{(a)} When predictions
are sorted into ten deciles by the size of the true step-to-step change $|\Delta y|$, CosDir improves
DA over the MSE baseline most on the largest moves, the ones that drive decisions. \textit{(b)}
Acting on the predicted direction, CosDir accumulates more cumulative payoff than MSE.}
\label{fig:mech}
\end{figure*}

\textbf{Comparison with various loss functions.}
Table~\ref{tab:sota} compares CosDir and CosDir-UW against shape,
frequency, and directional losses across all 15 backbones and 13 datasets, each added to the same
MSE base over matched cells. Our methods outperform all of them in DA, with CosDir-UW adding a
further gain over CosDir.
For fairness, every loss uses the \emph{same} $\lambda=0.5$, and
the per-$\lambda$ sweep (Appendix~\ref{app:lambda}) shows CosDir wins across the grid, with
CosDir-UW's \emph{learned} $\lambda$ matching a per-dataset \emph{tuned} $\lambda$ (Appendix~\ref{app:uw}).

\textbf{Ablation of the direction-aware term.}
To isolate what makes CosDir effective, namely its scale-invariant cosine formulation, we compare it
against \textit{three alternative direction-aware terms}, none of them
scale-invariant: 1) a first-difference MSE term, 2) a magnitude-weighted sign penalty, and 3) a per-step sign-classification (BCE) term
(see Appendix~\ref{app:ablvar} for details).
These are not existing losses but
variants that each inject directional
information in a different, non-scale-invariant way, so any gap to CosDir isolates the effect of
scale invariance. As shown in
Table~\ref{tab:ablation}, the CosDir family reaches the highest DA across all 15 backbones and 13 datasets. The
magnitude-weighted variants help less
as
their directional gradient still vanishes on small
moves, so the gain comes
from scale invariance rather than from adding a directional term.
\section{Analysis}

We analyze CosDir along five main axes, with further analyses deferred to the appendix:
\begin{itemize}[leftmargin=1.6em,itemsep=1pt,topsep=2pt]
\item (1) Where the directional gains come from
\item (2) Magnitude preservation across settings
\item (3) Cross-backbone consistency of the learned ratio
\item (4) From a fixed to an adaptive mixing ratio
\item (5) Robustness across backbones and datasets
\item (6) Other analyses (Appendix):
  \begin{itemize}[leftmargin=1.1em,itemsep=0pt,topsep=1pt,label=\textendash]
  \item Sensitivity to the lookback length (App.~\ref{app:robust})
  \item Generality across MSE/MAE base losses (App.~\ref{app:robust})
  \item Complementarity with the FreDF loss (App.~\ref{app:robust})
  \item Transaction-cost robustness of the payoff (App.~\ref{app:robustgain})
  \item Robustness of the gains to class balance (App.~\ref{app:robustgain})
  \item Significance testing and runtime (App.~\ref{app:protocol})
  \item Sensitivity to the weight $\lambda$ (App.~\ref{app:lambda})
  \end{itemize}
\end{itemize}

\begin{figure}[t]
\centering
\includegraphics[width=\columnwidth]{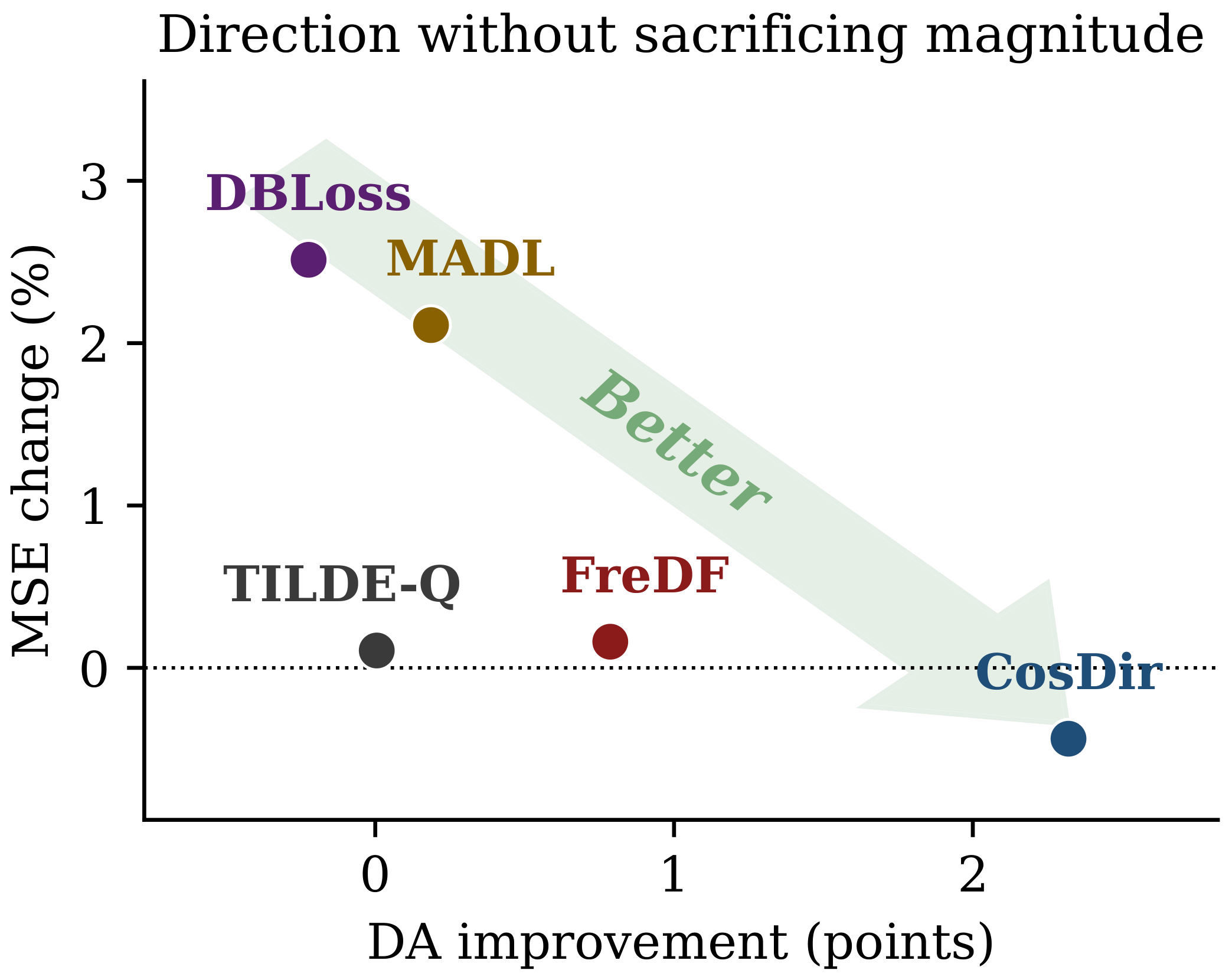}
\caption{Direction without sacrificing magnitude. On financial datasets, CosDir raises DA while
keeping or slightly improving MSE, whereas other auxiliary losses worsen MSE.}
\label{fig:mse}
\end{figure}

\begin{figure*}[t]
\centering
\includegraphics[width=\textwidth]{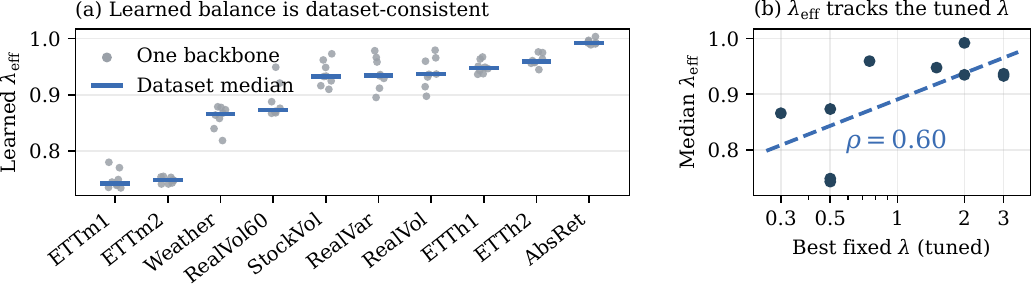}
\caption{CosDir-UW learns a dataset property. \textit{(a)} The learned balance
$\lambda_{\mathrm{eff}}=e^{\,s_1-s_2}$ clusters tightly by dataset across the eight main
backbones, with 96\% of its variance lying between datasets. \textit{(b)} The dataset-level
median $\lambda_{\mathrm{eff}}$ tracks the tuned $\lambda$.}
\label{fig:lameff}
\end{figure*}

\textbf{Where the gains come from.}
To trace where the directional gains come from, we group prediction points by the size of the true
step-to-step change $|\Delta y|$ on RealVol60 with PatchTST. Specifically, we sort them into ten
deciles and read DA within each. As shown in Figure~\ref{fig:mech}(a), both losses are near chance on the smallest moves, while
CosDir's advantage \textit{grows steadily with the size of the move}.
The gains therefore concentrate on the larger, decision-driving moves: CosDir does not
manufacture signal on the near-random smallest moves, but rather stops MSE from discarding the
signal the larger moves carry.
Figure~\ref{fig:mech}(b) translates this into payoff: \textit{acting on the predicted direction
accumulates more} with CosDir than with MSE.
These gains survive transaction costs and are not a class-imbalance artifact
(Appendix~\ref{app:robustgain}).

\textbf{Direction is improved without sacrificing magnitude.}
An auxiliary term is only useful if it does not degrade the magnitude accuracy it is added to.
Figure~\ref{fig:mse} places each
auxiliary loss by its DA gain and MSE change against the base loss
alone (w/o the auxiliary term), for FreTS on the five financial
datasets across horizons and seeds. CosDir sits at the lower right, raising DA without harming MSE,
whereas the other
losses worsen MSE. Full results are in Appendix~\ref{app:magnitude}.

\begin{table}[t]
\centering
\small
\setlength{\tabcolsep}{2.5pt}
\begin{tabular}{c|c|cc|cc|cc|cc}
\toprule
& \multirow{2.5}{*}{$\lambda$} & \multicolumn{2}{c|}{RealVar} & \multicolumn{2}{c|}{RealVol} & \multicolumn{2}{c|}{RealVol60} & \multicolumn{2}{c}{StockVol} \\
\cmidrule(lr){3-4}\cmidrule(lr){5-6}\cmidrule(lr){7-8}\cmidrule(lr){9-10}
& & DA & MSE & DA & MSE & DA & MSE & DA & MSE \\
\midrule
\multirow{10}{*}{\rotatebox{90}{\textbf{CosDir}}} & 0.05 & 51.5 & 0.463 & 51.4 & 0.471 & 51.3 & \underline{0.263} & 51.1 & 0.450 \\
& 0.1 & 51.7 & 0.463 & 51.6 & \underline{0.470} & 51.5 & 0.264 & 51.2 & \underline{0.448} \\
& 0.2 & 51.9 & \underline{0.462} & 51.7 & \underline{0.470} & 51.7 & 0.265 & 51.3 & 0.450 \\
& 0.3 & 52.0 & 0.464 & 51.9 & 0.471 & 51.9 & 0.267 & 51.5 & 0.450 \\
& 0.5 & 51.9 & 0.499 & 51.8 & 0.506 & \underline{52.3} & \textbf{0.261} & 51.4 & 0.473 \\
& 0.75 & 52.3 & 0.465 & 52.1 & 0.472 & \underline{52.3} & 0.269 & 51.7 & 0.452 \\
& 1.0 & 52.4 & 0.466 & \underline{52.2} & 0.473 & \underline{52.3} & 0.273 & \underline{51.8} & 0.452 \\
& 1.5 & 52.4 & 0.471 & \underline{52.2} & 0.476 & \underline{52.3} & 0.278 & \underline{51.8} & 0.454 \\
& 2.0 & \underline{52.5} & 0.473 & \textbf{52.3} & 0.479 & \textbf{52.4} & 0.282 & \textbf{51.9} & 0.459 \\
& 3.0 & \underline{52.5} & 0.482 & \textbf{52.3} & 0.489 & \underline{52.3} & 0.290 & \textbf{51.9} & 0.462 \\
\midrule
\multicolumn{2}{c|}{\textbf{CosDir-UW}} & \textbf{52.6} & \textbf{0.449} & \textbf{52.3} & \textbf{0.457} & \textbf{52.4} & \underline{0.263} & \textbf{51.9} & \textbf{0.440} \\
\bottomrule
\end{tabular}
\caption{
Learned $\lambda$ outperforms fixed $\lambda$. By balancing direction
and magnitude, CosDir-UW (learned $\lambda$) reaches a higher DA and lower MSE than CosDir
(fixed $\lambda$).
}
\label{tab:lam_ds}
\end{table}

\textbf{Cross-backbone consistency of the learned ratio.}
If CosDir-UW truly reads how much directional signal the data carries, the balance it learns
should depend on the dataset and not on the model trained with it. We train CosDir-UW on 10
datasets and the eight main backbones and record the learned
$\lambda_{\text{eff}}=e^{\,s_1-s_2}$ of every run. As shown in Figure~\ref{fig:lameff}(a), the
learned balance clusters tightly by dataset, and $96\%$ of the variance in
$\log\lambda_{\text{eff}}$ lies between datasets with only $4\%$ across backbones. As shown in
Figure~\ref{fig:lameff}(b), the dataset-level balance also tracks the tuned fixed $\lambda$
(Spearman $\rho=0.60$). The learned ratio is therefore a property of the data rather than a model
artifact, which is exactly what a hyperparameter-free replacement for a tuned $\lambda$ should
recover.

\textbf{From a fixed to an adaptive mixing ratio.}
The weight $\lambda$ sets the balance between direction and magnitude. By learning $\lambda$ rather
than fixing it, CosDir-UW improves on both DA (direction) and MSE (magnitude) at once. As Table~\ref{tab:lam_ds} shows, averaged over all 15 backbones on the
four financial datasets, the learned $\lambda$ gives a higher DA and a lower MSE than a fixed $\lambda$, with
no tuning. Appendix~\ref{app:uw} shows that the learned ratio tracks the per-dataset tuned $\lambda$.

\begin{figure}[t]
\centering
\includegraphics[width=\columnwidth]{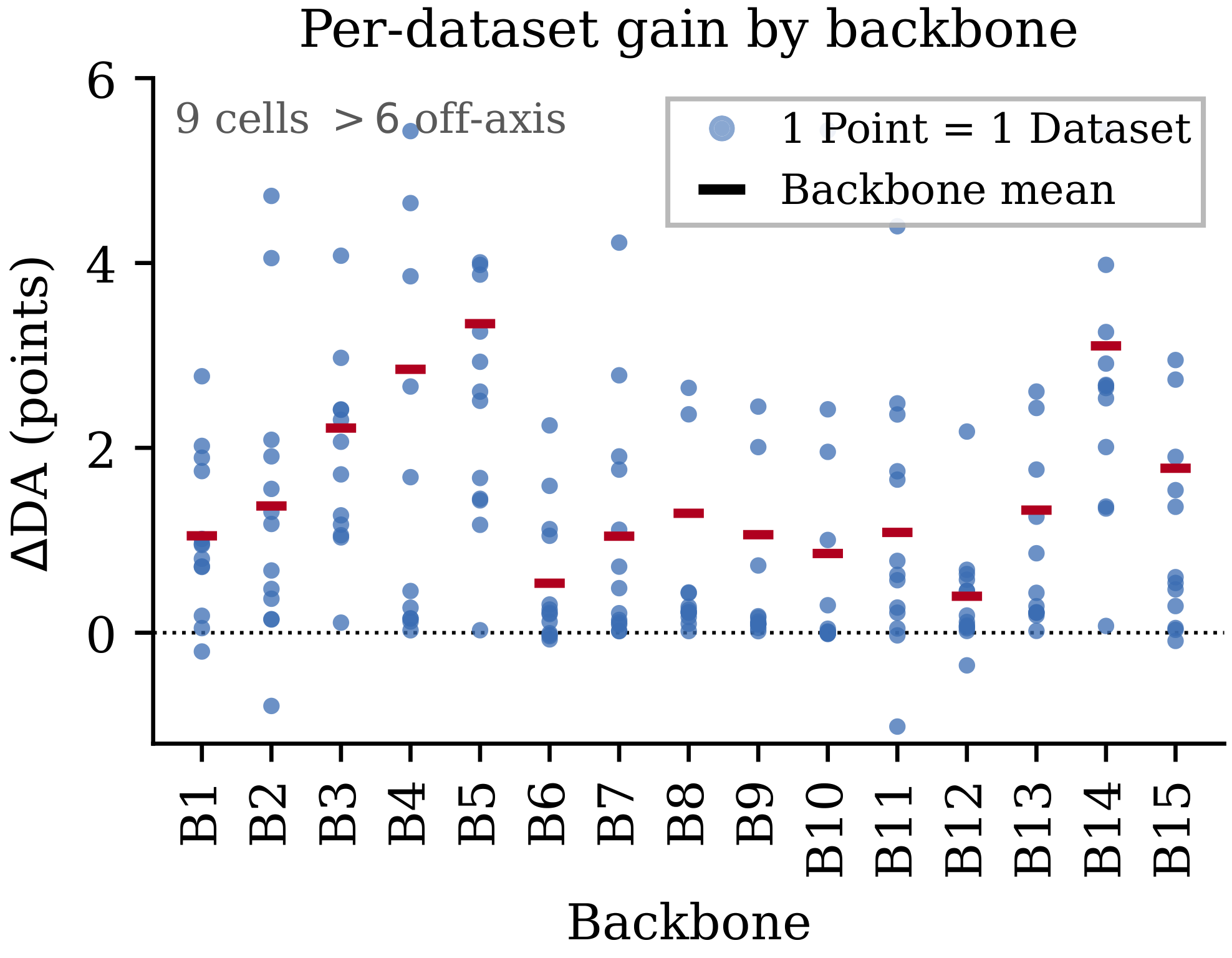}
\caption{Robustness across backbones and datasets. For each backbone (B1--B15), one point is the DA
gain of CosDir over MSE on one dataset, and the dash is the backbone mean.
}
\label{fig:robust}
\end{figure}

\textbf{Robustness across backbones and datasets.}
The directional gain is broad rather than tied to a particular model or domain. As
Figure~\ref{fig:robust} shows, across the 15 backbones and 13 datasets, adding CosDir improves DA on
$93\%$ of (backbone, dataset) cells,
while leaving MSE essentially
unchanged.
The backbone codes B1--B15 are listed in Appendix~\ref{app:backbones}.
\section{Conclusion}

We identify a failure mode of magnitude losses: dominated by large moves, MSE-trained forecasters
mispredict the direction of small changes, where most directional errors occur. We propose
\textbf{CosDir}, a scale-invariant loss aligning prediction and target difference vectors via
cosine similarity, and \textbf{CosDir-UW}, which learns the direction--magnitude balance. Over 100K+
experiments on 13 datasets and 15 backbones,
they consistently improve DA while
preserving magnitude. %

\textbf{Limitations and future works.}
CosDir targets the direction of change, so its gains are smaller on near-random returns, a
limit of the data, not the loss. Future directions include masking or
confidence-weighting near-flat windows and extending it to
probabilistic, multi-step forecasting.

\bibliography{aaai2027}

\clearpage
\setcounter{secnumdepth}{1}
\appendix
\section{Dataset Details}
\label{app:data}

The financial datasets are built entirely from publicly available daily price data
downloaded from Yahoo Finance over 2010--2024. Each channel is the risk measure of one individual
asset, so the number of channels $C$ equals the number of assets. From each asset's adjusted close
we compute daily log returns and then form rolling risk measures. Four datasets share a common panel
of $C{=}27$ large-cap U.S.\ equities: \textbf{RealVar} (log realized variance), \textbf{RealVol}
(log realized volatility over a 20-day window), \textbf{RealVol60} (the same over a 60-day window),
and \textbf{AbsRet} (rolling mean absolute return). \textbf{StockVol} is the per-asset realized
volatility for a separate basket of $C{=}24$ equities. All are constructed from open market data, and we will release the
construction code and the datasets to support reproducibility. The eight general benchmarks (ETT,
Weather, Solar, Traffic, Electricity) are standard public forecasting benchmarks~\citep{wu2021}.
All datasets use a chronological $70/10/20$ train/validation/test split.

\section{Backbone Details}
\label{app:backbones}

\noindent\textbf{Transformer-based.}
\begin{itemize}[leftmargin=*, itemsep=0pt, topsep=2pt]
  \item \textbf{[B2] PatchTST}~\citep{nie2023}: Channel-independent patching of the series fed to a Transformer encoder.
  \item \textbf{[B3] iTransformer}~\citep{liu2024}: Inverted attention where each variate (channel) is embedded as a token.
  \item \textbf{[B6] Autoformer}~\citep{wu2021}:     Decomposition with auto-correlation instead of dot-product attention.
  \item \textbf{[B7] FEDformer}~\citep{zhou2022}: Frequency-enhanced decomposed Transformer (Fourier basis).
  \item \textbf{[B10] Pyraformer}~\citep{liu2022pyraformer}: Pyramidal attention capturing multi-resolution dependencies at low cost.
  \item \textbf{[B11] Crossformer}~\citep{zhang2023crossformer}: Cross-dimension attention modeling inter-variable dependency.
\end{itemize}

\noindent\textbf{MLP-based.}
\begin{itemize}[leftmargin=*, itemsep=0pt, topsep=2pt]
  \item \textbf{[B1] DLinear}~\citep{zeng2023}: Trend--seasonal decomposition followed by a single linear layer per component.
  \item \textbf{[B5] TimeMixer}~\citep{wang2024timemixer}: Decomposable multiscale mixing of seasonal and trend representations.
  \item \textbf{[B8] TSMixer}~\citep{chen2023}: All-MLP architecture mixing alternately across time and channels.
  \item \textbf{[B9] LightTS}~\citep{zhang2022lightts}: Light sampling-oriented MLP over interval and continuous subsequences.
  \item \textbf{[B12] FiLM}~\citep{zhou2022film}: Legendre memory projections with frequency-enhanced low-rank filtering.
  \item \textbf{[B13] TiDE}~\citep{das2023tide}: Dense MLP encoder--decoder with covariate projection.
  \item \textbf{[B14] FreTS}~\citep{yi2023frets}: MLPs applied in the frequency domain over both channel and temporal axes.
\end{itemize}

\noindent\textbf{CNN/RNN-based.}
\begin{itemize}[leftmargin=*, itemsep=0pt, topsep=2pt]
  \item \textbf{[B4] TimesNet}~\citep{wu2023}: Reshapes the 1D series into 2D period-wise tensors processed by convolutions.
  \item \textbf{[B15] SegRNN}~\citep{lin2023segrnn}: Segment-wise GRU recurrence that replaces point-wise recurrent steps.
\end{itemize}

All 15 backbones use the default architecture hyperparameters of the Time-Series-Library
(MM-TSFlib), shared across models for a controlled comparison (Table~\ref{tab:model_hparams}).
\begin{table}[h]
\centering
\small
\setlength{\tabcolsep}{3.5pt}
\begin{tabular}{@{}l|ccccc|l@{}}
\toprule
Backbone & $d$ & $N_e$ & $N_d$ & $h$ & $d_{\mathrm{ff}}$ & Model-specific \\
\midrule
\multicolumn{7}{@{}l}{\textit{Transformer-based}} \\
PatchTST     & 128 & 2   & ---  & 8   & 256 & Factor $1$ \\
iTransformer & 128 & 2   & ---  & 8   & 256 & Factor $1$ \\
Autoformer   & 128 & 2   & 1    & 8   & 256 & $w{=}25$, factor $1$ \\
FEDformer    & 128 & 2   & 1    & 8   & 256 & $w{=}25$ \\
Pyraformer   & 128 & --- & ---  & --- & --- & --- \\
Crossformer  & 128 & 2   & ---  & 8   & 256 & Factor $1$ \\
\midrule
\multicolumn{7}{@{}l}{\textit{MLP-based}} \\
DLinear      & --- & --- & ---  & --- & --- & $w{=}25$ \\
TSMixer      & 128 & 2   & ---  & --- & --- & --- \\
TimeMixer    & 128 & 2   & ---  & --- & 256 & $w{=}25$, top-$k$ $5$ \\
TiDE         & 128 & 2   & 1    & --- & 256 & --- \\
LightTS      & 128 & --- & ---  & --- & --- & --- \\
FreTS        & --- & --- & ---  & --- & --- & Internal embedding \\
FiLM         & 128 & 2   & ---  & --- & --- & --- \\
\midrule
\multicolumn{7}{@{}l}{\textit{CNN-based}} \\
TimesNet     & 128 & 2   & ---  & --- & 256 & Top-$k$ $5$, kernels $6$ \\
\midrule
\multicolumn{7}{@{}l}{\textit{RNN-based}} \\
SegRNN       & 128 & --- & ---  & --- & --- & Seg-len $48$ \\
\bottomrule
\end{tabular}
\caption{Per-backbone hyperparameters, all following
default values of MM-TSFlib.
Dropout is $0.1$ throughout.
$d$: model dimension; $N_e$/$N_d$: encoder/decoder layers; $h$: attention heads; $d_{\mathrm{ff}}$:
FFN dimension; $w$: moving-average
window.}
\label{tab:model_hparams}
\end{table}

\section{Gradient of the Directional Term}
\label{app:grad}
We derive the gradient of the cosine penalty in Eq.~\eqref{eq:cosdir} to make precise
the sense in which CosDir keeps a directional signal on small moves. Fix a channel and write
$\mathbf{u}=\Delta\hat{\mathbf{y}}_c$ and $\mathbf{v}=\Delta\mathbf{y}_c$, with unit vectors
$\hat{\mathbf{u}}=\mathbf{u}/\lVert\mathbf{u}\rVert$ and
$\hat{\mathbf{v}}=\mathbf{v}/\lVert\mathbf{v}\rVert$. The per-channel loss is
$\ell=1-\cos(\mathbf{u},\mathbf{v})$ with
$\cos(\mathbf{u},\mathbf{v})=\langle\mathbf{u},\mathbf{v}\rangle/(\lVert\mathbf{u}\rVert\lVert\mathbf{v}\rVert)$
(we drop $\epsilon$ for clarity). Differentiating with respect to the prediction gives
\begin{equation}
\frac{\partial \ell}{\partial \mathbf{u}}
= -\frac{1}{\lVert\mathbf{u}\rVert}\Big(\hat{\mathbf{v}}-\cos(\mathbf{u},\mathbf{v})\,\hat{\mathbf{u}}\Big).
\label{eq:cosgrad}
\end{equation}
This gradient has two effects that explain why CosDir keeps pushing on direction even when
the moves are small.

The first effect is that it equalizes the directional signal across windows and channels. The
gradient carries a $1/\lVert\mathbf{u}\rVert$ prefactor, so a window whose overall move is small
(small $\lVert\mathbf{u}\rVert$) actually receives a \emph{larger} directional update rather than a
smaller one. This is the opposite of what happens with the MSE gradient
$\propto(\hat{\mathbf{y}}-\mathbf{y})$ and with a first-difference gradient
$\propto(\mathbf{u}-\mathbf{v})$, both of which shrink toward zero as the move shrinks. In practice
this means that MSE and first-difference terms quietly stop learning on exactly the low-amplitude
windows, whereas CosDir treats every window on an equal footing regardless of its amplitude. This is
why the small-amplitude series that magnitude losses tend to ignore are no longer left behind.

The second effect is that it controls how the update is spread over the individual steps inside a
single window. For horizon step $h$ the update is
$-\tfrac{1}{\lVert\mathbf{u}\rVert}\big(\hat v_h-\cos(\mathbf{u},\mathbf{v})\,\hat u_h\big)$, which
nudges the predicted step toward the unit target direction $\hat v_h$ and then discounts it by how
well the prediction is already aligned. What decides the contribution of a step is therefore the
normalized shape of the target $\hat{\mathbf{v}}$, not the raw size of its move. A small move inside a
window is scaled down only in proportion to $\hat v_h$, so it is dampened rather than erased
altogether. This is exactly the behavior we describe in the main text. The equalization is exact
across windows and channels, and inside a window the steps keep their ordering by normalized shape,
not by the raw size of the move.

\section{Pseudocode of CosDir}
\label{app:algo}
Algorithm~\ref{alg:cosdir} gives the forward pass of the CosDir and CosDir-UW losses as a
plug-in on top of any base loss. The only overhead beyond the base loss is the $O(HC)$
difference-and-cosine computation (lines 1--6), which leaves training and inference time essentially
unchanged (Appendix~\ref{app:protocol}).

\begin{algorithm}[h]
\caption{CosDir / CosDir-UW directional loss}
\label{alg:cosdir}
\begin{algorithmic}[1]
\Require prediction $\hat{Y}\in\mathbb{R}^{H\times C}$, target $Y\in\mathbb{R}^{H\times C}$, last input
$y_{\text{last}}\in\mathbb{R}^{1\times C}$; weight $\lambda$ (CosDir) or log-variances $s_1,s_2$
(CosDir-UW); stabilizer $\epsilon$
\State $\Delta\hat{Y} \gets \hat{Y} - [\,y_{\text{last}};\,\hat{Y}_{1:H-1}\,]$ \Comment{first differences along the horizon}
\State $\Delta Y \gets Y - [\,y_{\text{last}};\,Y_{1:H-1}\,]$
\For{$c = 1$ \textbf{to} $C$}
    \State $\cos_c \gets \dfrac{\langle \Delta\hat{Y}_{:,c},\, \Delta Y_{:,c}\rangle}{\lVert\Delta\hat{Y}_{:,c}\rVert\,\lVert\Delta Y_{:,c}\rVert + \epsilon}$
\EndFor
\State $\mathcal{L}_{\text{dir}} \gets \frac{1}{C}\sum_{c=1}^{C}(1 - \cos_c)$
\State $\mathcal{L}_{\text{MSE}} \gets \frac{1}{HC}\lVert \hat{Y} - Y\rVert_F^2$
\If{using CosDir}
    \State $\mathcal{L} \gets \mathcal{L}_{\text{MSE}} + \lambda\,\mathcal{L}_{\text{dir}}$
\Else \Comment{CosDir-UW: learn the balance via uncertainty weighting}
    \State $\mathcal{L} \gets e^{-s_1}\mathcal{L}_{\text{MSE}} + e^{-s_2}\mathcal{L}_{\text{dir}} + \tfrac{1}{2}(s_1 + s_2)$
\EndIf
\State backpropagate $\mathcal{L}$ and update the model (and $s_1, s_2$ for CosDir-UW)
\State \Return $\mathcal{L}$ \Comment{report $\lambda_{\text{eff}} = e^{\,s_1 - s_2}$ for CosDir-UW}
\end{algorithmic}
\end{algorithm}
\section{Direction-Aware Ablation Variants}
\label{app:ablvar}
The three non-scale-invariant terms in the ablation (Table~\ref{tab:ablation}) use the \emph{same}
first-difference vectors $\Delta\hat{\mathbf{y}}$ and $\Delta\mathbf{y}$, each added to the MSE base with weight $\lambda$.

\textbf{First-difference MSE.} A squared error on the differences,
\begin{equation}
\mathcal{L} = \mathcal{L}_{\mathrm{MSE}}
+ \frac{\lambda}{HC}\sum_{c,h}\big(\Delta\hat{y}_{h,c}-\Delta y_{h,c}\big)^2 .
\end{equation}
Its directional gradient is proportional to $\Delta\hat{y}-\Delta y$, so it shrinks with the true
move and vanishes on small moves.

\textbf{Magnitude-weighted sign penalty.} A smooth sign-agreement term, weighted by the true move
magnitude,
\begin{equation}
\mathcal{L} = \mathcal{L}_{\mathrm{MSE}}
+ \frac{\lambda}{HC}\sum_{c,h}
\lvert\Delta y_{h,c}\rvert\big(1-\tanh(k\,\Delta\hat{y}_{h,c}\,\Delta y_{h,c})\big),
\end{equation}
with $k=10$. The $\lvert\Delta y\rvert$ weight makes the directional signal fade exactly where moves
are small.

\textbf{Sign-classification (BCE).} A per-step binary cross-entropy that reads the predicted
change (scaled by $k$) as the logit for an ``up'' step,
\begin{equation}
\mathcal{L} = \mathcal{L}_{\mathrm{MSE}}
+ \frac{\lambda}{HC}\sum_{c,h}\mathrm{BCE}\!\big(k\,\Delta\hat{y}_{h,c},\, \mathbf{1}[\Delta y_{h,c}>0]\big),
\end{equation}
with $k=5$. This is the natural sign-classification auxiliary. It improves DA over the baseline but, unlike CosDir, it
optimizes a hard up/down label that ignores how strongly the trajectories agree (Table~\ref{tab:ablation}).

All three inject directional information in a non-scale-invariant way, unlike CosDir's scale-invariant cosine, so the gap to CosDir isolates the effect of scale invariance.

\section{Robustness and Sensitivity of CosDir}
\label{app:robust}
This section probes the robustness and sensitivity of CosDir along the four aspects below. Unless a
table notes otherwise, every number is the mean over the 5 random seeds.

\textbf{Sensitivity to $\lambda$.}
A fixed $\lambda$ is effective across its whole range, and CosDir improves DA at every $\lambda$.
Table~\ref{tab:lam_ds} in the main paper reports this on the financial datasets, and the full sweep
is in Appendix~\ref{app:lambda}. We use $\lambda=0.5$ by default and make the ratio adaptive in
CosDir-UW.

\textbf{Lookback length.}
Table~\ref{tab:lookback} varies the lookback $L$ on three datasets (RealVar, ETTh2, Solar) across the
eight main backbones\footnote{DLinear, PatchTST, iTransformer,
TimesNet, TimeMixer, Autoformer, FEDformer, and TSMixer.}. CosDir improves DA consistently, and the
improvement grows with larger $L$, confirming that the directional signal it captures is not
an artifact of a particular input length.

\begin{table}[h]
\centering
\small
\begin{tabular}{lccc}
\toprule
Lookback $L$ & MSE & $+$\,CosDir & $\Delta$DA \\
\midrule
96 & $0.5524_{\,\pm0.0004}$ & $\mathbf{0.5660}_{\,\pm0.0002}$ & +1.36 \\
192 & $0.6056_{\,\pm0.0009}$ & $\mathbf{0.6277}_{\,\pm0.0004}$ & +2.21 \\
336 & $0.6062_{\,\pm0.0017}$ & $\mathbf{0.6300}_{\,\pm0.0004}$ & +2.38 \\
\bottomrule
\end{tabular}%
\caption{Effect of the lookback length $L$. CosDir improves DA at every
lookback, with larger gains for longer histories.}
\label{tab:lookback}
\end{table}

\textbf{Base loss.}
Table~\ref{tab:baseloss} shows that CosDir is agnostic to the base loss on four datasets (RealVar,
RealVol, ETTh2, Solar) across the eight main backbones. Adding CosDir to either an
MSE or an MAE base improves DA,
so the directional term is complementary to the choice of point loss.

\begin{table}[h]
\centering
\small
\begin{tabular}{lccc}
\toprule
Base loss & --- & $+$\,CosDir & $\Delta$DA \\
\midrule
MSE & $0.5349_{\,\pm0.0003}$ & $\mathbf{0.5456}_{\,\pm0.0003}$ & +1.06 \\
MAE & $0.5805_{\,\pm0.0005}$ & $\mathbf{0.5906}_{\,\pm0.0004}$ & +1.01 \\
\bottomrule
\end{tabular}%
\caption{Base-loss ablation. CosDir improves DA on both an MSE and an MAE base loss.}
\label{tab:baseloss}
\end{table}

\textbf{Complementarity with a frequency loss.}
Table~\ref{tab:complement} combines CosDir with the frequency loss FreDF on four datasets (ETTh1,
ETTh2, Weather, RealVar) across the eight main backbones. The combination attains
the best DA, indicating that direction and frequency
are complementary axes and that CosDir complements existing structure-aware losses.%

\begin{table}[h]
\centering
\small
\begin{tabular}{lc}
\toprule
Loss & DA ($\uparrow$) \\
\midrule
MSE & $0.5330_{\,\pm0.0006}$ \\
+\,FreDF & $0.5426_{\,\pm0.0001}$ \\
+\,CosDir & $0.5461_{\,\pm0.0003}$ \\
+\,(FreDF \& CosDir) & $\mathbf{0.5696}_{\,\pm0.0004}$ \\
\bottomrule
\end{tabular}%
\caption{Complementarity with a frequency loss. Combining CosDir with FreDF attains the best DA, showing that direction and frequency are complementary.}
\label{tab:complement}
\end{table}

\section{Magnitude Preservation}
\label{app:magnitude}
Table~\ref{tab:mse_preserve} gives the numbers behind Figure~\ref{fig:mse}. On the five financial
datasets with FreTS~\citep{yi2023frets}, it reports the change in DA and MSE of each auxiliary loss relative to the base loss. CosDir and
CosDir-UW raise DA the most while keeping MSE at or below the baseline, whereas the shape and
frequency losses buy little direction and clearly worsen the very MSE they were meant to preserve.

\begin{table}[h]
\centering
\small
\begin{tabular}{lcc}
\toprule
Auxiliary loss & $\Delta$DA (pp) & $\Delta$MSE (\%) \\
\midrule
FreDF                     & $+0.79$ & $+0.16$ \\
TILDE-Q                   & $+0.00$ & $+0.11$ \\
DBLoss                    & $-0.22$ & $+2.51$ \\
MADL                      & $+0.19$ & $+2.11$ \\
\midrule
\textbf{CosDir (Ours)}    & $\underline{+2.32}$ & $\underline{-0.44}$ \\
\textbf{CosDir-UW (Ours)} & $\mathbf{+3.06}$ & $\mathbf{-0.56}$\\
\bottomrule
\end{tabular}
\caption{Magnitude preservation on the financial datasets. Change in DA and MSE per auxiliary
loss vs.\ the base loss. CosDir raises DA while keeping MSE below the baseline.}
\label{tab:mse_preserve}
\end{table}

\section{The Learned Balance in CosDir-UW}
\label{app:uw}

\textbf{Why an adaptive ratio: the best $\lambda$ is dataset-dependent.}
A fixed $\lambda$ already improves direction on every dataset, but the \textit{best} value differs.
Table~\ref{tab:bestlambda} lists, for the ten datasets used in the CosDir-UW analysis (Figure~\ref{fig:lameff}), the fixed $\lambda$ with the highest DA that still keeps MSE at the baseline, tuned per (dataset, backbone) over the eight main backbones, and it spans a $10\times$ range, from $0.3$ to $3.0$. CosDir-UW therefore learns this ratio per dataset rather than committing to a single fixed value shared uniformly by all datasets.

\begin{table}[h]
\centering
\small
\begin{tabular}{lc|lc}
\toprule
\multicolumn{2}{c|}{\textit{Financial}} & \multicolumn{2}{c}{\textit{General}} \\
\cmidrule(lr){1-2}\cmidrule(l){3-4}
Dataset & Best $\lambda$ & Dataset & Best $\lambda$ \\
\midrule
RealVar   & $2.0$ & ETTh1       & $1.5$ \\
RealVol   & $3.0$ & ETTh2       & $0.75$ \\
RealVol60 & $0.5$ & ETTm1       & $0.5$ \\
AbsRet    & $2.0$ & ETTm2       & $0.5$ \\
StockVol  & $3.0$ & Weather     & $0.3$ \\
\bottomrule
\end{tabular}
\caption{Best fixed $\lambda$ per dataset, the highest-DA value keeping MSE at or below baseline.
It spans an order of magnitude, motivating the learned balance in CosDir-UW.}
\label{tab:bestlambda}
\end{table}

\textbf{Per-dataset results for CosDir-UW.}
Table~\ref{tab:uwperdataset} reports absolute DA for the MSE baseline, CosDir at the default
$\lambda=0.5$, and the hyperparameter-free CosDir-UW, over all 15 backbones and every
horizon. CosDir-UW attains the best DA on all 13 datasets, with its clearest margin over a fixed $\lambda$ on the financial series, where the directionally optimal weight is furthest from the default.

\begin{table}[h]
\centering
\small
\begin{tabular}{ll ccc}
\toprule
Domain & Dataset & MSE & $+$\,CosDir & $+$\,CosDir-UW \\
\midrule
\multirow{8}{*}{\rotatebox{90}{General}} & ETTh1 & $0.5723$ & $0.5881$ & $\mathbf{0.5922}$ \\
 & ETTh2 & $0.5382$ & $0.5647$ & $\mathbf{0.5688}$ \\
 & ETTm1 & $0.5383$ & $0.5465$ & $\mathbf{0.5467}$ \\
 & ETTm2 & $0.5261$ & $0.5405$ & $\mathbf{0.5421}$ \\
 & Weather & $0.5239$ & $0.5411$ & $\mathbf{0.5439}$ \\
 & Solar & $0.6693$ & $0.7249$ & $\mathbf{0.7284}$ \\
 & Traffic & $0.7242$ & $0.7397$ & $\mathbf{0.7427}$ \\
 & Electricity & $0.6952$ & $0.7095$ & $\mathbf{0.7127}$ \\
\midrule
\multirow{5}{*}{\rotatebox{90}{Financial}} & RealVar & $0.5116$ & $0.5191$ & $\mathbf{0.5256}$ \\
 & RealVol & $0.5109$ & $0.5179$ & $\mathbf{0.5232}$ \\
 & RealVol60 & $0.5084$ & $0.5228$ & $\mathbf{0.5242}$ \\
 & AbsRet & $0.5077$ & $0.5082$ & $\mathbf{0.5083}$ \\
 & StockVol & $0.5084$ & $0.5140$ & $\mathbf{0.5189}$ \\
\bottomrule
\end{tabular}
\caption{Per-dataset DA for the MSE baseline, CosDir
($\lambda=0.5$), and the hyperparameter-free CosDir-UW, averaged over all 15 backbones and all
horizons.}
\label{tab:uwperdataset}
\end{table}
\section{Robustness of the Directional Gains}
\label{app:robustgain}

\textbf{Transaction costs.} To test whether the payoff survives realistic frictions, we
re-evaluate the directional strategy on RealVol60 with PatchTST under a turnover penalty. CosDir keeps
a higher net payoff than MSE at every penalty level, and its relative advantage grows with cost: from
$2.8\times$ at zero penalty to over $4\times$ once the penalty reaches a tenth of a typical move. Even
at a penalty of half a typical move, CosDir remains profitable while MSE has already turned negative.
The directional gain thus reflects fewer, better-timed reversals rather than churn.

\textbf{Class balance.} We check that the gains are not an artifact of class imbalance. On
RealVol60 with PatchTST the up/down split is nearly even ($49.5\%$ up), so DA is not inflated by a
majority label: an always-up predictor scores $0.495$, at chance. CosDir reaches $0.566$ against
$0.516$ for MSE, and the same ordering holds under balanced accuracy ($0.566$ vs.\ $0.516$) and the
Matthews correlation coefficient ($0.132$ vs.\ $0.032$, a $4\times$ increase). The improvement is thus a
genuine gain in up- and down-move recall alike, not a class shift.

\section{Significance Testing and Runtime}
\label{app:protocol}

\textbf{Significance threshold.} Given the many matched (dataset, backbone, horizon, seed)
cells, we use a conservative $p<10^{-3}$ threshold rather than $0.05$, a threshold that stays significant
under a Bonferroni correction over the cells reported in each table.

\textbf{Runtime.} Adding CosDir costs only the $O(HC)$ cosine computation on top of the forward
pass. In a controlled comparison under identical settings (iTransformer on Weather), training with
CosDir shows no measurable wall-clock overhead relative to the base loss ($26.7$ vs.\ $27.1$ s for three
epochs, within run-to-run variance), and inference is unchanged.

\textbf{Infrastructure.} All experiments were run on NVIDIA L40S GPUs (48\,GB) with PyTorch and the
Time-Series-Library (MM-TSFlib) under Linux.

\section{Per-Seed Standard Deviations}
\label{app:std}

For completeness, this appendix reproduces the main-body loss-comparison and ablation tables,
macro-averaged DA over the 13 datasets and 15 backbones, with the standard deviation
over the 5 random seeds (2024--2028) shown as a subscript on each mean. The means match those
reported in the main text, and the small standard deviations confirm that the effect of CosDir is
stable across seeds. Table~\ref{tab:sota_std} corresponds to Table~\ref{tab:sota}, and
Table~\ref{tab:ablation_std} to Table~\ref{tab:ablation}. The per-seed standard deviations of the
main DA results are small, so they are omitted from the horizon-resolved
Table~\ref{tab:main} for readability.

\begin{table}[h]
\centering
\small
\setlength{\tabcolsep}{3pt}
\begin{tabular}{lc}
\toprule
Loss & DA ($\uparrow$) \\
\midrule
MSE & $0.5642_{\,\pm0.0006}$ \\
+\,FreDF & $0.5729_{\,\pm0.0006}$ \\
+\,TILDE-Q & $0.5642_{\,\pm0.0004}$ \\
+\,DBLoss & $0.5618_{\,\pm0.0006}$ \\
+\,MADL & $0.5629_{\,\pm0.0003}$ \\
\textbf{+\,CosDir (Ours)} & $0.5798_{\,\pm0.0009}$ \\
\textbf{+\,CosDir-UW (Ours)} & $\mathbf{0.5829}_{\,\pm0.0007}$ \\
\bottomrule
\end{tabular}
\caption{CosDir vs.\ various loss functions.}
\label{tab:sota_std}
\end{table}

\begin{table}[h]
\centering
\small
\setlength{\tabcolsep}{3pt}
\begin{tabular}{lc}
\toprule
Loss & DA ($\uparrow$) \\
\midrule
MSE & $0.5642_{\,\pm0.0006}$ \\
+\,First-Difference & $0.5692_{\,\pm0.0004}$ \\
+\,Magnitude-Weighted Sign & $0.5663_{\,\pm0.0003}$ \\
+\,Sign Classification (BCE) & $0.5776_{\,\pm0.0011}$ \\
\textbf{+\,CosDir (Ours)} & $0.5798_{\,\pm0.0009}$ \\
\textbf{+\,CosDir-UW (Ours)} & $\mathbf{0.5829}_{\,\pm0.0007}$ \\
\bottomrule
\end{tabular}
\caption{Direction-aware loss ablation.}
\label{tab:ablation_std}
\end{table}
\section{Full Sensitivity to the Directional Weight $\lambda$}

\label{app:lambda}

This appendix reports the complete sensitivity of CosDir to its single mixing weight $\lambda$, which we fix to $0.5$ in all main experiments. The gain grows quickly up to $\lambda\!=\!0.5$ and then flattens, so $0.5$ is a strong default, and the \emph{highest-DA} $\lambda$ nonetheless differs by dataset (bold cells in Tables~\ref{tab:lamds}--\ref{tab:lampair12}), which is why we make the mixing ratio adaptive in CosDir-UW rather than rely on any single fixed value. Tables~\ref{tab:lamds}--\ref{tab:lambbmse} break the sweep down by dataset and by backbone, and Tables~\ref{tab:lampair0}--\ref{tab:lampair12} give the full per-(dataset$\times$backbone) grid. All cells are the mean over horizons and seeds, and the highest-DA $\lambda$ in each row is in bold.

\begin{table*}[t]
\centering
\small
\setlength{\tabcolsep}{4pt}
\begin{tabular}{l c | cccccccccc}
\toprule
 & MSE & \multicolumn{10}{c}{$\lambda$ (DA)} \\
\cmidrule(lr){3-12}
Group & Base & $0.05$ & $0.1$ & $0.2$ & $0.3$ & $0.5$ & $0.75$ & $1.0$ & $1.5$ & $2.0$ & $3.0$ \\
\midrule
ETTh1 & 0.572 & 0.576 & 0.579 & 0.583 & 0.586 & 0.588 & 0.591 & 0.592 & 0.594 & 0.594 & \textbf{0.595} \\
ETTh2 & 0.538 & 0.545 & 0.550 & 0.556 & 0.560 & 0.565 & 0.567 & 0.569 & 0.572 & 0.573 & \textbf{0.575} \\
ETTm1 & 0.538 & 0.540 & 0.542 & 0.544 & 0.545 & 0.546 & 0.547 & \textbf{0.547} & 0.547 & 0.547 & 0.546 \\
ETTm2 & 0.526 & 0.531 & 0.533 & 0.537 & 0.539 & 0.540 & 0.542 & 0.542 & 0.543 & 0.544 & \textbf{0.544} \\
Weather & 0.524 & 0.528 & 0.531 & 0.535 & 0.538 & 0.541 & 0.544 & 0.545 & 0.546 & 0.547 & \textbf{0.547} \\
Solar & 0.669 & 0.686 & 0.696 & 0.709 & 0.717 & 0.725 & 0.727 & 0.729 & 0.730 & 0.730 & \textbf{0.732} \\
Traffic & 0.724 & 0.731 & 0.734 & 0.738 & 0.741 & 0.740 & 0.748 & 0.749 & 0.751 & 0.752 & \textbf{0.752} \\
Electricity & 0.695 & 0.706 & 0.709 & 0.713 & 0.716 & 0.710 & 0.721 & 0.722 & 0.724 & 0.725 & \textbf{0.727} \\
RealVar & 0.512 & 0.515 & 0.517 & 0.519 & 0.520 & 0.519 & 0.523 & 0.524 & 0.524 & 0.525 & \textbf{0.525} \\
RealVol & 0.511 & 0.514 & 0.516 & 0.517 & 0.519 & 0.518 & 0.521 & 0.522 & 0.522 & 0.523 & \textbf{0.523} \\
RealVol60 & 0.508 & 0.513 & 0.515 & 0.517 & 0.519 & 0.523 & 0.523 & 0.524 & 0.525 & 0.525 & \textbf{0.526} \\
AbsRet & 0.508 & 0.508 & 0.508 & 0.508 & 0.508 & 0.508 & 0.508 & 0.508 & 0.508 & \textbf{0.508} & 0.508 \\
StockVol & 0.508 & 0.511 & 0.512 & 0.513 & 0.515 & 0.514 & 0.517 & 0.518 & 0.518 & 0.519 & \textbf{0.519} \\
\bottomrule
\end{tabular}
\caption{DA per dataset at each $\lambda$, averaged over the 15 backbones, horizons, and seeds, with the MSE baseline for reference.}
\label{tab:lamds}
\end{table*}

\begin{table*}[t]
\centering
\small
\setlength{\tabcolsep}{4pt}
\begin{tabular}{l c | cccccccccc}
\toprule
 & MSE & \multicolumn{10}{c}{$\lambda$ (MSE)} \\
\cmidrule(lr){3-12}
Group & Base & $0.05$ & $0.1$ & $0.2$ & $0.3$ & $0.5$ & $0.75$ & $1.0$ & $1.5$ & $2.0$ & $3.0$ \\
\midrule
ETTh1 & 0.554 & 0.553 & 0.553 & \textbf{0.552} & 0.553 & 0.556 & 0.556 & 0.558 & 0.563 & 0.568 & 0.576 \\
ETTh2 & 0.651 & 0.652 & 0.652 & 0.652 & 0.652 & \textbf{0.650} & 0.653 & 0.663 & 0.671 & 0.682 & 0.687 \\
ETTm1 & 0.465 & 0.467 & \textbf{0.467} & 0.468 & 0.469 & 0.468 & 0.472 & 0.474 & 0.481 & 0.488 & 0.499 \\
ETTm2 & 0.448 & 0.453 & 0.453 & 0.455 & 0.458 & \textbf{0.452} & 0.457 & 0.455 & 0.455 & 0.458 & 0.459 \\
Weather & 0.273 & \textbf{0.274} & 0.274 & 0.275 & 0.276 & 0.277 & 0.280 & 0.281 & 0.285 & 0.289 & 0.296 \\
Solar & 0.360 & \textbf{0.356} & 0.360 & 0.356 & 0.366 & 0.372 & 0.370 & 0.375 & 0.388 & 0.394 & 0.404 \\
Traffic & 0.781 & \textbf{0.752} & 0.752 & 0.755 & 0.757 & 0.786 & 0.763 & 0.768 & 0.777 & 0.784 & 0.798 \\
Electricity & 0.277 & 0.247 & 0.247 & \textbf{0.246} & 0.247 & 0.287 & 0.250 & 0.251 & 0.254 & 0.258 & 0.266 \\
RealVar & 0.496 & 0.463 & 0.463 & \textbf{0.462} & 0.464 & 0.499 & 0.465 & 0.466 & 0.471 & 0.473 & 0.482 \\
RealVol & 0.504 & 0.471 & 0.470 & \textbf{0.470} & 0.471 & 0.506 & 0.472 & 0.473 & 0.476 & 0.479 & 0.489 \\
RealVol60 & 0.255 & 0.263 & 0.264 & 0.265 & 0.267 & \textbf{0.261} & 0.269 & 0.273 & 0.278 & 0.282 & 0.290 \\
AbsRet & 0.867 & \textbf{0.869} & 0.869 & 0.870 & 0.871 & 0.872 & 0.874 & 0.876 & 0.878 & 0.881 & 0.886 \\
StockVol & 0.471 & 0.450 & \textbf{0.448} & 0.450 & 0.450 & 0.473 & 0.452 & 0.452 & 0.454 & 0.459 & 0.462 \\
\bottomrule
\end{tabular}
\caption{Magnitude error (MSE, lower is better) per dataset at each $\lambda$, averaged over the 15 backbones, horizons, and seeds.}
\label{tab:lamdsmse}
\end{table*}

\begin{table*}[t]
\centering
\small
\setlength{\tabcolsep}{4pt}
\begin{tabular}{l c | cccccccccc}
\toprule
 & MSE & \multicolumn{10}{c}{$\lambda$ (DA)} \\
\cmidrule(lr){3-12}
Group & Base & $0.05$ & $0.1$ & $0.2$ & $0.3$ & $0.5$ & $0.75$ & $1.0$ & $1.5$ & $2.0$ & $3.0$ \\
\midrule
Autoformer & 0.542 & 0.546 & 0.547 & 0.548 & 0.549 & 0.546 & 0.551 & 0.553 & 0.553 & 0.554 & \textbf{0.556} \\
Crossformer & 0.539 & 0.539 & 0.541 & 0.544 & 0.545 & 0.548 & 0.550 & 0.551 & 0.554 & 0.555 & \textbf{0.557} \\
DLinear & 0.566 & 0.573 & 0.576 & 0.578 & \textbf{0.578} & 0.576 & 0.576 & 0.576 & 0.575 & 0.574 & 0.573 \\
FEDformer & 0.551 & 0.556 & 0.557 & 0.560 & 0.561 & 0.559 & 0.565 & 0.566 & 0.567 & 0.567 & \textbf{0.568} \\
FiLM & 0.595 & 0.590 & 0.591 & 0.591 & 0.591 & \textbf{0.599} & 0.590 & 0.590 & 0.590 & 0.590 & 0.590 \\
FreTS & 0.553 & 0.564 & 0.571 & 0.578 & 0.581 & 0.582 & 0.588 & 0.589 & 0.590 & 0.591 & \textbf{0.592} \\
LightTS & 0.526 & 0.528 & 0.530 & 0.532 & 0.533 & 0.534 & 0.537 & 0.538 & 0.539 & 0.540 & \textbf{0.541} \\
PatchTST & 0.569 & 0.574 & 0.578 & 0.581 & 0.584 & 0.584 & 0.588 & \textbf{0.589} & 0.589 & 0.589 & 0.588 \\
Pyraformer & 0.545 & 0.547 & 0.548 & 0.549 & 0.550 & 0.551 & 0.551 & 0.551 & 0.552 & \textbf{0.552} & 0.551 \\
SegRNN & 0.545 & 0.550 & 0.552 & 0.555 & 0.558 & 0.559 & 0.562 & 0.563 & 0.563 & \textbf{0.563} & 0.563 \\
TSMixer & 0.531 & 0.534 & 0.536 & 0.538 & 0.541 & 0.541 & 0.545 & 0.547 & 0.548 & 0.549 & \textbf{0.551} \\
TiDE & 0.532 & 0.525 & 0.527 & 0.529 & 0.530 & \textbf{0.542} & 0.534 & 0.535 & 0.535 & 0.536 & 0.536 \\
TimeMixer & 0.546 & 0.531 & 0.537 & 0.546 & 0.551 & \textbf{0.575} & 0.558 & 0.559 & 0.560 & 0.560 & 0.561 \\
TimesNet & 0.558 & 0.566 & 0.571 & 0.575 & 0.579 & 0.579 & 0.584 & 0.586 & 0.587 & 0.588 & \textbf{0.589} \\
iTransformer & 0.553 & 0.559 & 0.563 & 0.567 & 0.571 & 0.573 & 0.578 & 0.580 & 0.582 & 0.583 & \textbf{0.584} \\
\bottomrule
\end{tabular}
\caption{DA per backbone at each $\lambda$, averaged over datasets, horizons, and seeds.}
\label{tab:lambb}
\end{table*}

\begin{table*}[t]
\centering
\small
\setlength{\tabcolsep}{4pt}
\begin{tabular}{l c | cccccccccc}
\toprule
 & MSE & \multicolumn{10}{c}{$\lambda$ (MSE)} \\
\cmidrule(lr){3-12}
Group & Base & $0.05$ & $0.1$ & $0.2$ & $0.3$ & $0.5$ & $0.75$ & $1.0$ & $1.5$ & $2.0$ & $3.0$ \\
\midrule
Autoformer & 0.601 & 0.588 & 0.588 & \textbf{0.580} & 0.591 & 0.611 & 0.589 & 0.593 & 0.601 & 0.604 & 0.612 \\
Crossformer & 0.566 & 0.510 & \textbf{0.510} & 0.511 & 0.512 & 0.575 & 0.512 & 0.519 & 0.526 & 0.528 & 0.526 \\
DLinear & 0.443 & 0.440 & \textbf{0.439} & 0.441 & 0.442 & 0.449 & 0.449 & 0.451 & 0.458 & 0.462 & 0.471 \\
FEDformer & 0.510 & \textbf{0.491} & 0.492 & 0.492 & 0.494 & 0.512 & 0.496 & 0.497 & 0.500 & 0.506 & 0.514 \\
FiLM & 0.453 & \textbf{0.429} & 0.429 & 0.431 & 0.432 & 0.460 & 0.435 & 0.436 & 0.439 & 0.440 & 0.443 \\
FreTS & 0.418 & \textbf{0.413} & 0.413 & 0.414 & 0.415 & 0.422 & 0.420 & 0.422 & 0.427 & 0.431 & 0.438 \\
LightTS & 0.492 & 0.481 & \textbf{0.481} & 0.481 & 0.481 & 0.494 & 0.484 & 0.486 & 0.491 & 0.496 & 0.506 \\
PatchTST & 0.420 & 0.413 & 0.413 & 0.413 & \textbf{0.412} & 0.419 & 0.415 & 0.416 & 0.421 & 0.424 & 0.433 \\
Pyraformer & 0.759 & \textbf{0.754} & 0.756 & 0.760 & 0.764 & 0.768 & 0.768 & 0.772 & 0.780 & 0.797 & 0.813 \\
SegRNN & 0.458 & 0.446 & \textbf{0.446} & 0.446 & 0.447 & 0.461 & 0.452 & 0.455 & 0.462 & 0.468 & 0.478 \\
TSMixer & 0.524 & 0.519 & 0.519 & \textbf{0.519} & 0.519 & 0.528 & 0.522 & 0.523 & 0.527 & 0.531 & 0.540 \\
TiDE & 0.464 & 0.443 & \textbf{0.443} & 0.443 & 0.443 & 0.466 & 0.446 & 0.447 & 0.451 & 0.454 & 0.460 \\
TimeMixer & 0.430 & \textbf{0.405} & 0.405 & 0.405 & 0.406 & 0.434 & 0.409 & 0.410 & 0.413 & 0.416 & 0.421 \\
TimesNet & 0.504 & \textbf{0.495} & 0.497 & 0.496 & 0.502 & 0.505 & 0.498 & 0.497 & 0.497 & 0.500 & 0.504 \\
iTransformer & 0.434 & 0.428 & 0.427 & 0.427 & \textbf{0.427} & 0.435 & 0.429 & 0.429 & 0.433 & 0.436 & 0.443 \\
\bottomrule
\end{tabular}
\caption{Magnitude error per backbone at each $\lambda$.}
\label{tab:lambbmse}
\end{table*}

\begin{table*}[t]
\centering
\small
\setlength{\tabcolsep}{4pt}
\begin{tabular}{l c | cccccccccc}
\toprule
 & MSE & \multicolumn{10}{c}{$\lambda$ (DA)} \\
\cmidrule(lr){3-12}
Group & Base & $0.05$ & $0.1$ & $0.2$ & $0.3$ & $0.5$ & $0.75$ & $1.0$ & $1.5$ & $2.0$ & $3.0$ \\
\midrule
Autoformer & 0.584 & 0.586 & 0.587 & 0.592 & 0.594 & 0.600 & 0.600 & 0.608 & 0.609 & 0.609 & \textbf{0.616} \\
Crossformer & 0.592 & 0.597 & 0.602 & 0.607 & 0.610 & 0.616 & 0.621 & 0.623 & 0.626 & 0.627 & \textbf{0.628} \\
DLinear & 0.582 & 0.590 & 0.593 & 0.597 & 0.599 & 0.602 & 0.603 & 0.604 & 0.605 & \textbf{0.606} & 0.606 \\
FEDformer & 0.568 & 0.569 & 0.573 & 0.579 & 0.582 & 0.585 & 0.590 & 0.592 & 0.595 & 0.596 & \textbf{0.598} \\
FiLM & 0.630 & 0.628 & 0.627 & 0.626 & 0.626 & 0.626 & 0.626 & 0.627 & 0.628 & 0.629 & \textbf{0.630} \\
FreTS & 0.591 & 0.601 & 0.607 & 0.612 & 0.616 & 0.618 & 0.622 & 0.624 & 0.626 & 0.627 & \textbf{0.628} \\
LightTS & 0.515 & 0.515 & 0.515 & 0.516 & 0.516 & 0.516 & 0.517 & 0.517 & 0.518 & \textbf{0.518} & 0.518 \\
PatchTST & 0.622 & 0.629 & 0.634 & 0.639 & 0.641 & 0.643 & 0.644 & 0.644 & 0.645 & 0.645 & \textbf{0.645} \\
Pyraformer & 0.526 & 0.528 & 0.529 & 0.530 & 0.531 & 0.529 & 0.531 & 0.531 & 0.531 & 0.532 & \textbf{0.533} \\
SegRNN & 0.555 & 0.560 & 0.564 & 0.569 & 0.572 & 0.574 & 0.578 & 0.579 & 0.581 & 0.582 & \textbf{0.583} \\
TSMixer & 0.524 & 0.525 & 0.525 & 0.527 & 0.528 & 0.529 & 0.531 & 0.532 & 0.533 & 0.534 & \textbf{0.535} \\
TiDE & 0.535 & 0.536 & 0.537 & 0.538 & 0.539 & 0.539 & 0.541 & \textbf{0.541} & 0.541 & 0.541 & 0.540 \\
TimeMixer & 0.565 & 0.569 & 0.572 & 0.577 & 0.579 & 0.579 & 0.584 & 0.584 & \textbf{0.584} & 0.584 & 0.583 \\
TimesNet & 0.610 & 0.618 & 0.631 & 0.641 & 0.648 & 0.657 & 0.662 & 0.664 & 0.666 & \textbf{0.667} & 0.667 \\
iTransformer & 0.586 & 0.591 & 0.595 & 0.601 & 0.605 & 0.609 & 0.612 & 0.614 & 0.616 & 0.617 & \textbf{0.618} \\
\bottomrule
\end{tabular}
\caption{Directional accuracy on ETTh1 for every backbone at each $\lambda$.}
\label{tab:lampair0}
\end{table*}

\begin{table*}[t]
\centering
\small
\setlength{\tabcolsep}{4pt}
\begin{tabular}{l c | cccccccccc}
\toprule
 & MSE & \multicolumn{10}{c}{$\lambda$ (DA)} \\
\cmidrule(lr){3-12}
Group & Base & $0.05$ & $0.1$ & $0.2$ & $0.3$ & $0.5$ & $0.75$ & $1.0$ & $1.5$ & $2.0$ & $3.0$ \\
\midrule
Autoformer & 0.572 & 0.579 & 0.585 & 0.590 & 0.592 & 0.595 & 0.598 & 0.602 & 0.602 & 0.601 & \textbf{0.605} \\
Crossformer & 0.513 & 0.516 & 0.518 & 0.522 & 0.525 & 0.529 & 0.533 & 0.550 & 0.579 & 0.588 & \textbf{0.594} \\
DLinear & 0.558 & 0.572 & 0.582 & \textbf{0.585} & 0.582 & 0.577 & 0.567 & 0.562 & 0.556 & 0.553 & 0.550 \\
FEDformer & 0.528 & 0.538 & 0.543 & 0.549 & 0.554 & 0.556 & 0.566 & 0.570 & 0.575 & 0.578 & \textbf{0.583} \\
FiLM & 0.631 & 0.633 & 0.634 & 0.634 & 0.635 & 0.635 & 0.636 & 0.636 & 0.636 & 0.636 & \textbf{0.636} \\
FreTS & 0.538 & 0.551 & 0.561 & 0.576 & 0.584 & 0.592 & 0.596 & 0.600 & 0.604 & 0.606 & \textbf{0.608} \\
LightTS & 0.505 & 0.505 & 0.505 & 0.505 & 0.506 & 0.506 & 0.507 & 0.507 & 0.507 & 0.508 & \textbf{0.508} \\
PatchTST & 0.582 & 0.593 & 0.601 & 0.613 & 0.617 & 0.622 & 0.623 & 0.624 & \textbf{0.624} & 0.624 & 0.623 \\
Pyraformer & 0.500 & 0.501 & 0.501 & 0.501 & 0.501 & 0.501 & 0.501 & 0.502 & 0.502 & 0.503 & \textbf{0.507} \\
SegRNN & 0.521 & 0.527 & 0.531 & 0.538 & 0.542 & 0.549 & 0.551 & 0.553 & 0.555 & 0.556 & \textbf{0.557} \\
TSMixer & 0.502 & 0.501 & 0.502 & 0.502 & 0.502 & 0.504 & 0.505 & 0.505 & 0.506 & 0.508 & \textbf{0.511} \\
TiDE & 0.514 & 0.515 & 0.516 & 0.518 & 0.520 & 0.523 & 0.524 & 0.525 & \textbf{0.525} & 0.525 & 0.524 \\
TimeMixer & 0.529 & 0.536 & 0.541 & 0.547 & 0.551 & 0.554 & \textbf{0.556} & 0.556 & 0.554 & 0.552 & 0.550 \\
TimesNet & 0.534 & 0.549 & 0.571 & 0.593 & 0.617 & 0.641 & 0.650 & 0.655 & 0.662 & 0.664 & \textbf{0.667} \\
iTransformer & 0.546 & 0.554 & 0.561 & 0.571 & 0.578 & 0.587 & 0.592 & 0.595 & 0.598 & 0.600 & \textbf{0.601} \\
\bottomrule
\end{tabular}
\caption{Directional accuracy on ETTh2 for every backbone at each $\lambda$.}
\label{tab:lampair1}
\end{table*}

\begin{table*}[t]
\centering
\small
\setlength{\tabcolsep}{4pt}
\begin{tabular}{l c | cccccccccc}
\toprule
 & MSE & \multicolumn{10}{c}{$\lambda$ (DA)} \\
\cmidrule(lr){3-12}
Group & Base & $0.05$ & $0.1$ & $0.2$ & $0.3$ & $0.5$ & $0.75$ & $1.0$ & $1.5$ & $2.0$ & $3.0$ \\
\midrule
Autoformer & 0.522 & 0.521 & 0.522 & 0.523 & 0.523 & 0.524 & 0.527 & 0.531 & 0.533 & 0.535 & \textbf{0.537} \\
Crossformer & 0.538 & 0.540 & 0.542 & 0.551 & 0.555 & 0.563 & 0.565 & 0.566 & \textbf{0.566} & 0.566 & 0.565 \\
DLinear & 0.578 & \textbf{0.581} & 0.581 & 0.579 & 0.578 & 0.576 & 0.575 & 0.574 & 0.573 & 0.573 & 0.572 \\
FEDformer & 0.529 & 0.530 & 0.529 & 0.529 & 0.530 & 0.531 & 0.531 & 0.531 & 0.531 & 0.531 & \textbf{0.532} \\
FiLM & 0.583 & 0.584 & \textbf{0.584} & 0.584 & 0.584 & 0.584 & 0.584 & 0.584 & 0.583 & 0.583 & 0.583 \\
FreTS & 0.529 & 0.532 & 0.535 & 0.539 & 0.540 & 0.542 & 0.543 & 0.544 & 0.544 & \textbf{0.544} & 0.544 \\
LightTS & 0.515 & 0.515 & 0.515 & 0.515 & 0.516 & \textbf{0.516} & 0.516 & 0.515 & 0.515 & 0.514 & 0.513 \\
PatchTST & 0.560 & 0.562 & 0.564 & 0.565 & 0.566 & \textbf{0.567} & 0.566 & 0.566 & 0.564 & 0.563 & 0.562 \\
Pyraformer & 0.514 & 0.514 & 0.514 & \textbf{0.515} & 0.514 & 0.514 & 0.513 & 0.513 & 0.512 & 0.511 & 0.510 \\
SegRNN & 0.534 & 0.536 & 0.537 & 0.538 & 0.539 & \textbf{0.540} & 0.540 & 0.539 & 0.538 & 0.537 & 0.535 \\
TSMixer & 0.519 & 0.519 & 0.519 & 0.520 & \textbf{0.520} & 0.520 & 0.519 & 0.518 & 0.517 & 0.516 & 0.515 \\
TiDE & 0.527 & 0.528 & 0.529 & \textbf{0.529} & 0.529 & 0.529 & 0.528 & 0.527 & 0.526 & 0.525 & 0.524 \\
TimeMixer & 0.530 & 0.534 & 0.539 & 0.549 & 0.555 & 0.559 & \textbf{0.560} & 0.560 & 0.557 & 0.555 & 0.551 \\
TimesNet & 0.555 & 0.563 & 0.569 & 0.575 & 0.578 & 0.581 & 0.583 & 0.586 & 0.590 & 0.590 & \textbf{0.592} \\
iTransformer & 0.542 & 0.545 & 0.547 & 0.550 & 0.551 & 0.553 & 0.553 & 0.553 & \textbf{0.553} & 0.553 & 0.552 \\
\bottomrule
\end{tabular}
\caption{Directional accuracy on ETTm1 for every backbone at each $\lambda$.}
\label{tab:lampair2}
\end{table*}

\begin{table*}[t]
\centering
\small
\setlength{\tabcolsep}{4pt}
\begin{tabular}{l c | cccccccccc}
\toprule
 & MSE & \multicolumn{10}{c}{$\lambda$ (DA)} \\
\cmidrule(lr){3-12}
Group & Base & $0.05$ & $0.1$ & $0.2$ & $0.3$ & $0.5$ & $0.75$ & $1.0$ & $1.5$ & $2.0$ & $3.0$ \\
\midrule
Autoformer & 0.546 & 0.546 & 0.549 & 0.555 & 0.554 & 0.557 & 0.558 & 0.558 & 0.557 & 0.560 & \textbf{0.562} \\
Crossformer & 0.503 & 0.504 & 0.506 & 0.506 & 0.507 & 0.511 & 0.516 & 0.518 & 0.521 & 0.526 & \textbf{0.531} \\
DLinear & 0.567 & 0.575 & 0.574 & \textbf{0.576} & 0.576 & 0.574 & 0.573 & 0.572 & 0.570 & 0.568 & 0.565 \\
FEDformer & 0.536 & 0.541 & 0.543 & 0.545 & 0.545 & 0.543 & 0.545 & 0.543 & 0.546 & 0.549 & \textbf{0.553} \\
FiLM & 0.581 & 0.585 & 0.585 & 0.586 & 0.587 & 0.587 & 0.588 & 0.588 & 0.588 & 0.588 & \textbf{0.588} \\
FreTS & 0.513 & 0.518 & 0.520 & 0.522 & 0.524 & 0.526 & 0.530 & 0.531 & 0.532 & 0.533 & \textbf{0.533} \\
LightTS & 0.504 & 0.504 & 0.504 & 0.504 & 0.505 & 0.506 & 0.506 & 0.506 & \textbf{0.506} & 0.506 & 0.505 \\
PatchTST & 0.544 & 0.550 & 0.552 & 0.555 & 0.556 & \textbf{0.556} & 0.556 & 0.555 & 0.555 & 0.555 & 0.554 \\
Pyraformer & 0.501 & \textbf{0.501} & 0.501 & 0.501 & 0.501 & 0.501 & 0.501 & 0.501 & 0.501 & 0.501 & 0.501 \\
SegRNN & 0.516 & 0.519 & \textbf{0.519} & 0.518 & 0.516 & 0.515 & 0.514 & 0.513 & 0.513 & 0.513 & 0.513 \\
TSMixer & 0.504 & 0.504 & 0.504 & 0.505 & 0.506 & 0.506 & 0.507 & 0.507 & \textbf{0.508} & 0.507 & 0.507 \\
TiDE & 0.520 & 0.527 & 0.531 & 0.535 & 0.537 & 0.538 & \textbf{0.538} & 0.537 & 0.536 & 0.535 & 0.532 \\
TimeMixer & 0.514 & 0.519 & 0.525 & 0.537 & 0.546 & 0.554 & 0.557 & 0.558 & \textbf{0.559} & 0.559 & 0.558 \\
TimesNet & 0.528 & 0.551 & 0.566 & 0.582 & 0.592 & 0.597 & 0.602 & 0.605 & 0.607 & 0.609 & \textbf{0.610} \\
iTransformer & 0.514 & 0.516 & 0.519 & 0.524 & 0.527 & 0.535 & 0.538 & 0.541 & 0.544 & 0.546 & \textbf{0.548} \\
\bottomrule
\end{tabular}
\caption{Directional accuracy on ETTm2 for every backbone at each $\lambda$.}
\label{tab:lampair3}
\end{table*}

\begin{table*}[t]
\centering
\small
\setlength{\tabcolsep}{4pt}
\begin{tabular}{l c | cccccccccc}
\toprule
 & MSE & \multicolumn{10}{c}{$\lambda$ (DA)} \\
\cmidrule(lr){3-12}
Group & Base & $0.05$ & $0.1$ & $0.2$ & $0.3$ & $0.5$ & $0.75$ & $1.0$ & $1.5$ & $2.0$ & $3.0$ \\
\midrule
Autoformer & 0.521 & 0.524 & 0.526 & 0.526 & 0.526 & 0.531 & 0.536 & 0.536 & 0.539 & 0.539 & \textbf{0.542} \\
Crossformer & 0.516 & 0.519 & 0.521 & 0.526 & 0.530 & 0.533 & 0.539 & 0.542 & 0.545 & 0.547 & \textbf{0.547} \\
DLinear & 0.558 & 0.565 & \textbf{0.567} & 0.567 & 0.566 & 0.565 & 0.565 & 0.564 & 0.563 & 0.563 & 0.561 \\
FEDformer & 0.529 & 0.534 & 0.537 & 0.542 & 0.546 & 0.548 & 0.554 & 0.555 & 0.556 & 0.556 & \textbf{0.557} \\
FiLM & 0.569 & 0.570 & 0.570 & 0.570 & 0.570 & 0.571 & 0.571 & 0.571 & 0.571 & 0.571 & \textbf{0.571} \\
FreTS & 0.513 & 0.525 & 0.533 & 0.541 & \textbf{0.542} & 0.540 & 0.540 & 0.538 & 0.538 & 0.538 & 0.539 \\
LightTS & 0.508 & 0.509 & 0.511 & 0.513 & 0.514 & 0.515 & 0.516 & 0.516 & 0.517 & 0.518 & \textbf{0.518} \\
PatchTST & 0.531 & 0.531 & 0.532 & 0.533 & 0.534 & 0.536 & 0.537 & 0.538 & 0.540 & 0.542 & \textbf{0.544} \\
Pyraformer & 0.508 & 0.509 & 0.510 & 0.514 & 0.518 & 0.528 & 0.534 & 0.538 & 0.543 & 0.546 & \textbf{0.549} \\
SegRNN & 0.511 & 0.513 & 0.516 & 0.519 & 0.522 & 0.527 & 0.530 & \textbf{0.531} & 0.530 & 0.528 & 0.524 \\
TSMixer & 0.511 & 0.511 & 0.512 & 0.513 & 0.514 & 0.515 & 0.516 & 0.518 & 0.519 & 0.521 & \textbf{0.523} \\
TiDE & 0.517 & 0.522 & 0.525 & 0.528 & 0.529 & 0.530 & 0.531 & 0.531 & 0.531 & 0.532 & \textbf{0.532} \\
TimeMixer & 0.515 & 0.519 & 0.523 & 0.533 & 0.542 & 0.555 & 0.560 & 0.562 & 0.564 & 0.564 & \textbf{0.564} \\
TimesNet & 0.527 & 0.544 & 0.558 & 0.572 & 0.577 & 0.581 & 0.586 & 0.585 & 0.587 & \textbf{0.588} & 0.587 \\
iTransformer & 0.523 & 0.525 & 0.528 & 0.532 & 0.535 & 0.540 & 0.544 & 0.547 & 0.550 & 0.552 & \textbf{0.553} \\
\bottomrule
\end{tabular}
\caption{Directional accuracy on Weather for every backbone at each $\lambda$.}
\label{tab:lampair4}
\end{table*}

\begin{table*}[t]
\centering
\small
\setlength{\tabcolsep}{4pt}
\begin{tabular}{l c | cccccccccc}
\toprule
 & MSE & \multicolumn{10}{c}{$\lambda$ (DA)} \\
\cmidrule(lr){3-12}
Group & Base & $0.05$ & $0.1$ & $0.2$ & $0.3$ & $0.5$ & $0.75$ & $1.0$ & $1.5$ & $2.0$ & $3.0$ \\
\midrule
Autoformer & 0.535 & 0.530 & 0.534 & 0.533 & 0.540 & 0.537 & 0.532 & 0.540 & 0.539 & 0.539 & \textbf{0.553} \\
Crossformer & 0.653 & 0.664 & 0.670 & 0.678 & 0.681 & \textbf{0.697} & 0.681 & 0.671 & 0.662 & 0.657 & 0.658 \\
DLinear & 0.749 & 0.750 & 0.750 & \textbf{0.750} & 0.750 & 0.750 & 0.749 & 0.749 & 0.748 & 0.747 & 0.746 \\
FEDformer & 0.697 & 0.705 & 0.713 & 0.723 & 0.728 & 0.740 & 0.744 & 0.749 & 0.754 & 0.753 & \textbf{0.758} \\
FiLM & 0.753 & 0.753 & 0.753 & 0.753 & 0.753 & \textbf{0.754} & 0.753 & 0.753 & 0.753 & 0.753 & 0.753 \\
FreTS & 0.654 & 0.702 & 0.727 & 0.742 & 0.747 & 0.749 & 0.750 & 0.752 & 0.753 & 0.753 & \textbf{0.753} \\
LightTS & 0.609 & 0.625 & 0.637 & 0.657 & 0.670 & 0.686 & 0.697 & 0.704 & 0.710 & 0.713 & \textbf{0.716} \\
PatchTST & 0.736 & 0.738 & 0.738 & 0.739 & 0.739 & 0.740 & 0.741 & \textbf{0.741} & 0.741 & 0.740 & 0.736 \\
Pyraformer & 0.720 & 0.757 & 0.768 & 0.773 & 0.774 & \textbf{0.774} & 0.773 & 0.772 & 0.772 & 0.769 & 0.762 \\
SegRNN & 0.643 & 0.657 & 0.671 & 0.699 & 0.724 & 0.750 & 0.759 & 0.762 & 0.764 & 0.765 & \textbf{0.766} \\
TSMixer & 0.627 & 0.642 & 0.656 & 0.681 & 0.700 & 0.722 & 0.735 & 0.742 & 0.749 & 0.752 & \textbf{0.755} \\
TiDE & 0.601 & 0.613 & 0.623 & 0.640 & 0.652 & 0.669 & 0.680 & 0.687 & 0.695 & 0.699 & \textbf{0.703} \\
TimeMixer & 0.619 & 0.658 & 0.697 & 0.745 & 0.758 & 0.765 & 0.766 & 0.767 & 0.767 & 0.767 & \textbf{0.767} \\
TimesNet & 0.730 & 0.756 & 0.761 & 0.764 & 0.766 & 0.769 & 0.770 & 0.771 & 0.772 & 0.773 & \textbf{0.774} \\
iTransformer & 0.710 & 0.735 & 0.747 & 0.759 & 0.766 & 0.771 & 0.773 & 0.774 & \textbf{0.774} & 0.774 & 0.773 \\
\bottomrule
\end{tabular}
\caption{Directional accuracy on Solar for every backbone at each $\lambda$.}
\label{tab:lampair5}
\end{table*}

\begin{table*}[t]
\centering
\small
\setlength{\tabcolsep}{4pt}
\begin{tabular}{l c | cccccccccc}
\toprule
 & MSE & \multicolumn{10}{c}{$\lambda$ (DA)} \\
\cmidrule(lr){3-12}
Group & Base & $0.05$ & $0.1$ & $0.2$ & $0.3$ & $0.5$ & $0.75$ & $1.0$ & $1.5$ & $2.0$ & $3.0$ \\
\midrule
Autoformer & 0.708 & 0.728 & 0.731 & 0.729 & 0.733 & 0.710 & 0.746 & 0.747 & 0.750 & 0.754 & \textbf{0.758} \\
Crossformer & 0.627 & 0.626 & 0.627 & 0.628 & 0.628 & \textbf{0.630} & 0.628 & 0.628 & 0.628 & 0.629 & 0.629 \\
DLinear & 0.771 & 0.776 & 0.778 & 0.779 & 0.780 & \textbf{0.780} & 0.780 & 0.779 & 0.779 & 0.779 & 0.779 \\
FEDformer & 0.738 & 0.752 & 0.754 & 0.758 & 0.760 & 0.750 & 0.767 & 0.769 & 0.772 & 0.774 & \textbf{0.775} \\
FreTS & 0.727 & 0.735 & 0.741 & 0.748 & 0.753 & 0.759 & 0.763 & 0.765 & 0.767 & 0.767 & \textbf{0.768} \\
LightTS & 0.623 & 0.625 & 0.627 & 0.632 & 0.636 & 0.643 & 0.649 & 0.655 & 0.663 & 0.668 & \textbf{0.675} \\
PatchTST & 0.770 & \textbf{0.769} & 0.768 & 0.767 & 0.765 & 0.762 & 0.757 & 0.753 & 0.746 & 0.740 & 0.731 \\
Pyraformer & 0.747 & 0.754 & 0.758 & 0.764 & 0.767 & 0.771 & \textbf{0.772} & 0.772 & 0.771 & 0.770 & 0.769 \\
SegRNN & 0.717 & 0.722 & 0.727 & 0.734 & 0.739 & 0.746 & 0.751 & 0.754 & 0.757 & 0.758 & \textbf{0.759} \\
TSMixer & 0.653 & 0.657 & 0.661 & 0.667 & 0.672 & 0.680 & 0.686 & 0.690 & 0.694 & 0.697 & \textbf{0.700} \\
TimesNet & 0.799 & 0.816 & 0.817 & 0.819 & 0.820 & 0.804 & 0.823 & 0.824 & 0.824 & 0.826 & \textbf{0.826} \\
iTransformer & 0.729 & 0.735 & 0.741 & 0.748 & 0.753 & 0.758 & 0.761 & 0.763 & 0.764 & 0.765 & \textbf{0.766} \\
\bottomrule
\end{tabular}
\caption{Directional accuracy on Traffic for every backbone at each $\lambda$.}
\label{tab:lampair6}
\end{table*}

\begin{table*}[t]
\centering
\small
\setlength{\tabcolsep}{4pt}
\begin{tabular}{l c | cccccccccc}
\toprule
 & MSE & \multicolumn{10}{c}{$\lambda$ (DA)} \\
\cmidrule(lr){3-12}
Group & Base & $0.05$ & $0.1$ & $0.2$ & $0.3$ & $0.5$ & $0.75$ & $1.0$ & $1.5$ & $2.0$ & $3.0$ \\
\midrule
Autoformer & 0.678 & 0.702 & 0.703 & 0.707 & 0.707 & 0.681 & 0.712 & 0.715 & 0.716 & \textbf{0.721} & 0.721 \\
Crossformer & 0.664 & 0.733 & 0.736 & 0.741 & 0.744 & 0.654 & 0.750 & 0.751 & 0.754 & 0.755 & \textbf{0.757} \\
DLinear & 0.708 & 0.712 & 0.715 & 0.719 & 0.722 & 0.726 & 0.728 & 0.731 & 0.734 & 0.737 & \textbf{0.742} \\
FEDformer & 0.681 & 0.700 & 0.704 & 0.709 & 0.712 & 0.686 & 0.720 & 0.722 & 0.724 & 0.725 & \textbf{0.726} \\
FiLM & 0.738 & 0.744 & 0.745 & 0.745 & 0.745 & 0.739 & 0.745 & 0.745 & \textbf{0.745} & 0.745 & 0.745 \\
FreTS & 0.733 & 0.744 & 0.751 & 0.756 & 0.758 & 0.760 & 0.761 & 0.761 & 0.762 & 0.762 & \textbf{0.762} \\
LightTS & 0.646 & 0.650 & 0.653 & 0.659 & 0.663 & 0.671 & 0.676 & 0.681 & 0.687 & 0.690 & \textbf{0.694} \\
PatchTST & 0.738 & 0.742 & 0.746 & \textbf{0.748} & 0.746 & 0.739 & 0.730 & 0.723 & 0.713 & 0.707 & 0.698 \\
Pyraformer & 0.687 & 0.696 & 0.699 & 0.704 & 0.706 & 0.697 & 0.708 & 0.710 & 0.711 & 0.712 & \textbf{0.714} \\
SegRNN & 0.690 & 0.693 & 0.695 & 0.698 & 0.700 & 0.704 & 0.707 & 0.709 & 0.713 & 0.715 & \textbf{0.717} \\
TSMixer & 0.668 & 0.672 & 0.675 & 0.680 & 0.685 & 0.692 & 0.698 & 0.703 & 0.709 & 0.713 & \textbf{0.717} \\
TiDE & 0.668 & 0.659 & 0.662 & 0.668 & 0.672 & 0.694 & 0.684 & 0.688 & 0.693 & 0.696 & \textbf{0.700} \\
TimesNet & 0.707 & 0.726 & 0.728 & 0.732 & 0.733 & 0.724 & 0.736 & 0.737 & 0.737 & 0.738 & \textbf{0.738} \\
iTransformer & 0.704 & 0.708 & 0.712 & 0.718 & 0.722 & 0.728 & 0.733 & 0.736 & 0.740 & 0.742 & \textbf{0.745} \\
\bottomrule
\end{tabular}
\caption{Directional accuracy on Electricity for every backbone at each $\lambda$.}
\label{tab:lampair7}
\end{table*}

\begin{table*}[t]
\centering
\small
\setlength{\tabcolsep}{4pt}
\begin{tabular}{l c | cccccccccc}
\toprule
 & MSE & \multicolumn{10}{c}{$\lambda$ (DA)} \\
\cmidrule(lr){3-12}
Group & Base & $0.05$ & $0.1$ & $0.2$ & $0.3$ & $0.5$ & $0.75$ & $1.0$ & $1.5$ & $2.0$ & $3.0$ \\
\midrule
Autoformer & 0.505 & 0.505 & \textbf{0.506} & 0.506 & 0.505 & 0.505 & 0.505 & 0.504 & 0.504 & 0.504 & 0.505 \\
Crossformer & 0.510 & 0.512 & 0.513 & 0.515 & 0.516 & 0.516 & 0.521 & 0.523 & 0.525 & 0.527 & \textbf{0.529} \\
DLinear & 0.513 & 0.518 & 0.522 & 0.523 & 0.525 & 0.523 & 0.525 & \textbf{0.526} & 0.525 & 0.525 & 0.523 \\
FEDformer & 0.507 & 0.508 & 0.508 & \textbf{0.508} & 0.508 & 0.507 & 0.508 & 0.508 & 0.508 & 0.508 & 0.508 \\
FiLM & 0.546 & 0.559 & \textbf{0.559} & 0.559 & 0.558 & 0.551 & 0.557 & 0.556 & 0.555 & 0.555 & 0.554 \\
FreTS & 0.517 & 0.529 & 0.536 & 0.545 & 0.549 & 0.546 & 0.555 & 0.557 & 0.557 & 0.558 & \textbf{0.559} \\
LightTS & 0.502 & 0.503 & 0.503 & 0.503 & 0.504 & 0.503 & 0.504 & 0.504 & 0.505 & 0.505 & \textbf{0.505} \\
PatchTST & 0.522 & 0.526 & 0.531 & 0.535 & 0.540 & 0.541 & 0.547 & 0.550 & 0.550 & \textbf{0.553} & 0.551 \\
Pyraformer & 0.502 & 0.502 & 0.502 & 0.502 & 0.502 & \textbf{0.502} & 0.502 & 0.502 & 0.502 & 0.502 & 0.502 \\
SegRNN & 0.513 & 0.517 & 0.518 & 0.520 & 0.520 & 0.518 & \textbf{0.521} & 0.520 & 0.520 & 0.520 & 0.519 \\
TSMixer & 0.504 & 0.506 & 0.506 & 0.507 & 0.508 & 0.507 & 0.510 & 0.511 & 0.511 & 0.512 & \textbf{0.513} \\
TiDE & 0.502 & 0.503 & 0.503 & 0.504 & 0.504 & 0.504 & 0.506 & 0.506 & 0.506 & 0.507 & \textbf{0.507} \\
TimeMixer & 0.512 & 0.520 & 0.524 & 0.529 & 0.533 & 0.527 & 0.541 & 0.543 & 0.545 & 0.546 & \textbf{0.547} \\
TimesNet & 0.504 & 0.504 & 0.504 & 0.504 & 0.505 & 0.505 & 0.506 & 0.508 & 0.510 & 0.511 & \textbf{0.513} \\
iTransformer & 0.514 & 0.518 & 0.521 & 0.524 & 0.527 & 0.527 & 0.533 & 0.535 & 0.536 & \textbf{0.538} & 0.538 \\
\bottomrule
\end{tabular}
\caption{Directional accuracy on RealVar for every backbone at each $\lambda$.}
\label{tab:lampair8}
\end{table*}

\begin{table*}[t]
\centering
\small
\setlength{\tabcolsep}{4pt}
\begin{tabular}{l c | cccccccccc}
\toprule
 & MSE & \multicolumn{10}{c}{$\lambda$ (DA)} \\
\cmidrule(lr){3-12}
Group & Base & $0.05$ & $0.1$ & $0.2$ & $0.3$ & $0.5$ & $0.75$ & $1.0$ & $1.5$ & $2.0$ & $3.0$ \\
\midrule
Autoformer & 0.505 & 0.507 & 0.506 & 0.506 & \textbf{0.507} & 0.505 & 0.505 & 0.506 & 0.505 & 0.505 & 0.505 \\
Crossformer & 0.510 & 0.512 & 0.512 & 0.515 & 0.516 & 0.515 & 0.520 & 0.521 & 0.523 & 0.524 & \textbf{0.526} \\
DLinear & 0.512 & 0.518 & 0.522 & 0.522 & 0.524 & 0.522 & 0.523 & \textbf{0.525} & 0.523 & 0.523 & 0.521 \\
FEDformer & 0.507 & 0.508 & 0.508 & 0.508 & 0.508 & 0.508 & 0.508 & 0.508 & 0.508 & 0.508 & \textbf{0.508} \\
FiLM & 0.541 & 0.552 & \textbf{0.553} & 0.552 & 0.552 & 0.546 & 0.551 & 0.551 & 0.550 & 0.550 & 0.549 \\
FreTS & 0.516 & 0.526 & 0.533 & 0.540 & 0.544 & 0.541 & 0.549 & 0.550 & 0.551 & 0.551 & \textbf{0.552} \\
LightTS & 0.502 & 0.503 & 0.503 & 0.503 & 0.503 & 0.503 & 0.504 & 0.504 & 0.505 & 0.505 & \textbf{0.505} \\
PatchTST & 0.520 & 0.524 & 0.528 & 0.531 & 0.535 & 0.536 & 0.541 & 0.544 & 0.544 & \textbf{0.546} & 0.545 \\
Pyraformer & 0.502 & 0.502 & 0.502 & 0.502 & 0.502 & \textbf{0.502} & 0.502 & 0.502 & 0.502 & 0.502 & 0.502 \\
SegRNN & 0.512 & 0.516 & 0.517 & 0.518 & 0.519 & 0.517 & \textbf{0.519} & 0.519 & 0.518 & 0.518 & 0.518 \\
TSMixer & 0.504 & 0.506 & 0.506 & 0.507 & 0.507 & 0.507 & 0.509 & 0.510 & 0.511 & 0.511 & \textbf{0.512} \\
TiDE & 0.502 & 0.503 & 0.503 & 0.504 & 0.504 & 0.505 & 0.505 & 0.506 & 0.506 & 0.506 & \textbf{0.507} \\
TimeMixer & 0.511 & 0.518 & 0.521 & 0.526 & 0.530 & 0.528 & 0.536 & 0.538 & 0.539 & 0.540 & \textbf{0.541} \\
TimesNet & 0.503 & 0.504 & 0.504 & 0.504 & 0.505 & 0.505 & 0.505 & 0.506 & 0.508 & 0.511 & \textbf{0.514} \\
iTransformer & 0.513 & 0.517 & 0.519 & 0.521 & 0.524 & 0.525 & 0.529 & 0.531 & 0.532 & \textbf{0.534} & 0.533 \\
\bottomrule
\end{tabular}
\caption{Directional accuracy on RealVol for every backbone at each $\lambda$.}
\label{tab:lampair9}
\end{table*}

\begin{table*}[t]
\centering
\small
\setlength{\tabcolsep}{4pt}
\begin{tabular}{l c | cccccccccc}
\toprule
 & MSE & \multicolumn{10}{c}{$\lambda$ (DA)} \\
\cmidrule(lr){3-12}
Group & Base & $0.05$ & $0.1$ & $0.2$ & $0.3$ & $0.5$ & $0.75$ & $1.0$ & $1.5$ & $2.0$ & $3.0$ \\
\midrule
Autoformer & 0.501 & 0.502 & 0.501 & 0.501 & 0.501 & 0.502 & 0.502 & 0.501 & 0.502 & 0.502 & \textbf{0.502} \\
Crossformer & 0.503 & 0.503 & 0.503 & 0.504 & 0.504 & 0.504 & 0.504 & 0.504 & 0.504 & 0.505 & \textbf{0.505} \\
DLinear & 0.512 & 0.526 & 0.533 & 0.537 & 0.536 & \textbf{0.540} & 0.528 & 0.524 & 0.520 & 0.518 & 0.516 \\
FEDformer & 0.503 & 0.504 & 0.505 & 0.505 & \textbf{0.505} & 0.505 & 0.504 & 0.504 & 0.503 & 0.502 & 0.502 \\
FiLM & 0.556 & 0.577 & 0.577 & 0.577 & 0.577 & \textbf{0.578} & 0.576 & 0.576 & 0.576 & 0.575 & 0.575 \\
FreTS & 0.508 & 0.515 & 0.520 & 0.532 & 0.538 & 0.548 & 0.554 & 0.558 & 0.561 & 0.564 & \textbf{0.566} \\
LightTS & 0.501 & 0.501 & 0.502 & 0.502 & 0.502 & 0.502 & 0.502 & 0.502 & 0.503 & 0.503 & \textbf{0.503} \\
PatchTST & 0.517 & 0.524 & 0.530 & 0.541 & 0.550 & 0.564 & 0.573 & 0.578 & 0.583 & 0.585 & \textbf{0.587} \\
Pyraformer & 0.500 & 0.500 & 0.500 & 0.500 & 0.500 & \textbf{0.500} & 0.500 & 0.500 & 0.500 & 0.500 & 0.500 \\
SegRNN & 0.504 & 0.505 & \textbf{0.505} & 0.505 & 0.505 & 0.505 & 0.504 & 0.504 & 0.504 & 0.503 & 0.503 \\
TSMixer & 0.502 & 0.502 & 0.502 & 0.503 & 0.503 & 0.504 & 0.504 & 0.505 & 0.505 & 0.506 & \textbf{0.507} \\
TiDE & 0.501 & 0.501 & 0.501 & 0.502 & 0.502 & 0.503 & 0.503 & 0.503 & 0.504 & 0.504 & \textbf{0.504} \\
TimeMixer & 0.509 & 0.515 & 0.521 & 0.528 & 0.534 & 0.547 & 0.545 & 0.547 & 0.550 & 0.552 & \textbf{0.554} \\
TimesNet & 0.500 & 0.500 & 0.501 & 0.501 & 0.502 & 0.503 & 0.503 & 0.504 & 0.505 & 0.505 & \textbf{0.506} \\
iTransformer & 0.508 & 0.513 & 0.517 & 0.523 & 0.528 & 0.532 & 0.541 & 0.545 & 0.549 & 0.552 & \textbf{0.555} \\
\bottomrule
\end{tabular}
\caption{Directional accuracy on RealVol60 for every backbone at each $\lambda$.}
\label{tab:lampair10}
\end{table*}

\begin{table*}[t]
\centering
\small
\setlength{\tabcolsep}{4pt}
\begin{tabular}{l c | cccccccccc}
\toprule
 & MSE & \multicolumn{10}{c}{$\lambda$ (DA)} \\
\cmidrule(lr){3-12}
Group & Base & $0.05$ & $0.1$ & $0.2$ & $0.3$ & $0.5$ & $0.75$ & $1.0$ & $1.5$ & $2.0$ & $3.0$ \\
\midrule
Autoformer & 0.509 & 0.510 & 0.510 & 0.510 & \textbf{0.511} & 0.509 & 0.510 & 0.510 & 0.510 & 0.511 & 0.511 \\
Crossformer & 0.508 & 0.508 & 0.507 & 0.507 & 0.507 & \textbf{0.508} & 0.508 & 0.507 & 0.507 & 0.507 & 0.507 \\
DLinear & 0.508 & 0.508 & 0.508 & 0.508 & 0.509 & 0.509 & 0.510 & 0.510 & 0.510 & \textbf{0.510} & 0.510 \\
FEDformer & 0.510 & 0.509 & 0.509 & 0.509 & 0.509 & \textbf{0.510} & 0.509 & 0.509 & 0.508 & 0.509 & 0.509 \\
FiLM & 0.508 & 0.508 & 0.509 & 0.509 & 0.509 & \textbf{0.510} & 0.509 & 0.509 & 0.509 & 0.509 & 0.509 \\
FreTS & 0.508 & 0.508 & 0.508 & 0.508 & 0.508 & 0.509 & 0.509 & 0.509 & 0.509 & 0.509 & \textbf{0.509} \\
LightTS & 0.507 & 0.507 & 0.507 & 0.507 & 0.507 & \textbf{0.507} & 0.507 & 0.507 & 0.507 & 0.507 & 0.507 \\
PatchTST & 0.507 & 0.507 & 0.507 & 0.508 & 0.508 & 0.508 & 0.509 & 0.509 & 0.510 & 0.510 & \textbf{0.510} \\
Pyraformer & 0.507 & 0.507 & 0.507 & 0.507 & 0.507 & \textbf{0.507} & 0.507 & 0.507 & 0.507 & 0.507 & 0.506 \\
SegRNN & 0.508 & 0.507 & 0.508 & 0.508 & 0.508 & \textbf{0.508} & 0.508 & 0.508 & 0.508 & 0.508 & 0.508 \\
TSMixer & 0.507 & 0.507 & 0.507 & 0.507 & 0.507 & 0.507 & 0.507 & 0.507 & 0.508 & 0.508 & \textbf{0.508} \\
TiDE & 0.507 & 0.507 & 0.507 & 0.507 & 0.507 & \textbf{0.508} & 0.507 & 0.507 & 0.507 & 0.507 & 0.507 \\
TimeMixer & 0.508 & 0.507 & 0.507 & 0.508 & 0.508 & \textbf{0.508} & 0.508 & 0.508 & 0.508 & 0.508 & 0.508 \\
TimesNet & 0.507 & 0.507 & 0.507 & 0.507 & 0.508 & 0.508 & 0.508 & 0.508 & 0.508 & \textbf{0.509} & 0.508 \\
iTransformer & 0.506 & 0.506 & 0.506 & 0.507 & 0.507 & 0.507 & 0.507 & 0.507 & 0.507 & \textbf{0.508} & 0.507 \\
\bottomrule
\end{tabular}
\caption{Directional accuracy on AbsRet for every backbone at each $\lambda$.}
\label{tab:lampair11}
\end{table*}

\begin{table*}[t]
\centering
\small
\setlength{\tabcolsep}{4pt}
\begin{tabular}{l c | cccccccccc}
\toprule
 & MSE & \multicolumn{10}{c}{$\lambda$ (DA)} \\
\cmidrule(lr){3-12}
Group & Base & $0.05$ & $0.1$ & $0.2$ & $0.3$ & $0.5$ & $0.75$ & $1.0$ & $1.5$ & $2.0$ & $3.0$ \\
\midrule
Autoformer & 0.506 & 0.506 & 0.506 & 0.506 & 0.506 & 0.506 & 0.506 & 0.506 & 0.506 & 0.506 & \textbf{0.507} \\
Crossformer & 0.506 & 0.507 & 0.508 & 0.508 & 0.508 & 0.508 & 0.511 & 0.512 & 0.513 & 0.515 & \textbf{0.516} \\
DLinear & 0.509 & 0.513 & 0.516 & 0.518 & \textbf{0.519} & 0.517 & 0.517 & 0.518 & 0.516 & 0.516 & 0.515 \\
FEDformer & 0.506 & 0.507 & 0.508 & 0.508 & 0.508 & 0.507 & 0.509 & \textbf{0.509} & 0.509 & 0.509 & 0.509 \\
FiLM & 0.535 & 0.546 & 0.547 & \textbf{0.547} & 0.547 & 0.542 & 0.547 & 0.547 & 0.546 & 0.546 & 0.546 \\
FreTS & 0.511 & 0.520 & 0.524 & 0.529 & 0.532 & 0.532 & 0.540 & 0.542 & 0.544 & 0.545 & \textbf{0.546} \\
LightTS & 0.501 & 0.501 & 0.502 & 0.502 & 0.502 & 0.502 & 0.502 & 0.502 & 0.502 & 0.503 & \textbf{0.503} \\
PatchTST & 0.514 & 0.517 & 0.519 & 0.522 & 0.526 & 0.528 & 0.534 & 0.538 & 0.540 & \textbf{0.543} & 0.542 \\
Pyraformer & 0.501 & 0.501 & 0.501 & 0.501 & 0.501 & 0.501 & \textbf{0.501} & 0.501 & 0.501 & 0.501 & 0.501 \\
SegRNN & 0.510 & 0.513 & 0.514 & 0.514 & \textbf{0.514} & 0.513 & 0.514 & 0.514 & 0.513 & 0.513 & 0.512 \\
TSMixer & 0.502 & 0.504 & 0.504 & 0.505 & 0.505 & 0.504 & 0.506 & 0.507 & 0.508 & 0.508 & \textbf{0.509} \\
TiDE & 0.501 & 0.502 & 0.502 & 0.502 & 0.503 & 0.504 & 0.504 & 0.504 & 0.505 & 0.505 & \textbf{0.505} \\
TimeMixer & 0.508 & 0.512 & 0.515 & 0.518 & 0.521 & 0.520 & 0.527 & 0.528 & 0.531 & 0.532 & \textbf{0.533} \\
TimesNet & 0.503 & 0.504 & 0.504 & 0.505 & 0.505 & 0.505 & 0.506 & 0.509 & 0.509 & \textbf{0.510} & 0.509 \\
iTransformer & 0.510 & 0.513 & 0.515 & 0.518 & 0.520 & 0.520 & 0.524 & 0.527 & 0.528 & 0.530 & \textbf{0.530} \\
\bottomrule
\end{tabular}
\caption{Directional accuracy on StockVol for every backbone at each $\lambda$.}
\label{tab:lampair12}
\end{table*}
\section{Per-Horizon Results with Std.}

For full transparency, we report the per-horizon directional accuracy of MSE\,$+$\,CosDir on
every dataset, one table per dataset (Tables~\ref{tab:detailstd_ETTh1}--\ref{tab:detailstd_sabsret}),
with the standard deviation over random seeds shown as a subscript on every mean. As in
the main text, bold marks cells where adding CosDir improves DA over the MSE baseline. The small standard deviations confirm stability across seeds.

\begin{table*}[t]
\centering
\small
\setlength{\tabcolsep}{2pt}
\begin{tabular}{lcccc}
\toprule
Backbone & $H=96$ & $H=192$ & $H=336$ & $H=720$ \\
\midrule
DLinear & $\mathbf{0.613}_{\pm0.00}$ & $\mathbf{0.606}_{\pm0.00}$ & $\mathbf{0.598}_{\pm0.00}$ & $\mathbf{0.590}_{\pm0.00}$ \\
PatchTST & $\mathbf{0.657}_{\pm0.00}$ & $\mathbf{0.649}_{\pm0.00}$ & $\mathbf{0.638}_{\pm0.00}$ & $\mathbf{0.628}_{\pm0.00}$ \\
iTransformer & $\mathbf{0.632}_{\pm0.00}$ & $\mathbf{0.615}_{\pm0.00}$ & $\mathbf{0.601}_{\pm0.00}$ & $\mathbf{0.587}_{\pm0.00}$ \\
TimesNet & $\mathbf{0.669}_{\pm0.00}$ & $\mathbf{0.663}_{\pm0.01}$ & $\mathbf{0.656}_{\pm0.01}$ & $\mathbf{0.641}_{\pm0.01}$ \\
TimeMixer & $\mathbf{0.606}_{\pm0.01}$ & $\mathbf{0.585}_{\pm0.01}$ & $\mathbf{0.579}_{\pm0.00}$ & $\mathbf{0.557}_{\pm0.01}$ \\
Autoformer & $\mathbf{0.610}_{\pm0.02}$ & $\mathbf{0.605}_{\pm0.00}$ & $\mathbf{0.593}_{\pm0.01}$ & $\mathbf{0.591}_{\pm0.02}$ \\
FEDformer & $\mathbf{0.609}_{\pm0.01}$ & $\mathbf{0.594}_{\pm0.02}$ & $\mathbf{0.583}_{\pm0.01}$ & $\mathbf{0.560}_{\pm0.01}$ \\
TSMixer & $\mathbf{0.542}_{\pm0.00}$ & $\mathbf{0.532}_{\pm0.00}$ & $\mathbf{0.526}_{\pm0.00}$ & $\mathbf{0.519}_{\pm0.00}$ \\
LightTS & $\mathbf{0.520}_{\pm0.00}$ & $\mathbf{0.517}_{\pm0.00}$ & $\mathbf{0.516}_{\pm0.00}$ & $\mathbf{0.514}_{\pm0.00}$ \\
Pyraformer & $\mathbf{0.539}_{\pm0.00}$ & $\mathbf{0.532}_{\pm0.00}$ & $\mathbf{0.528}_{\pm0.00}$ & $\mathbf{0.523}_{\pm0.00}$ \\
Crossformer & $\mathbf{0.638}_{\pm0.00}$ & $\mathbf{0.628}_{\pm0.00}$ & $\mathbf{0.618}_{\pm0.00}$ & $\mathbf{0.584}_{\pm0.00}$ \\
FiLM & $0.655_{\pm0.00}$ & $0.640_{\pm0.00}$ & $0.617_{\pm0.00}$ & $0.595_{\pm0.00}$ \\
TiDE & $\mathbf{0.547}_{\pm0.00}$ & $\mathbf{0.542}_{\pm0.00}$ & $\mathbf{0.538}_{\pm0.00}$ & $\mathbf{0.534}_{\pm0.00}$ \\
FreTS & $\mathbf{0.640}_{\pm0.00}$ & $\mathbf{0.625}_{\pm0.00}$ & $\mathbf{0.611}_{\pm0.00}$ & $\mathbf{0.601}_{\pm0.00}$ \\
SegRNN & $\mathbf{0.588}_{\pm0.00}$ & $\mathbf{0.581}_{\pm0.00}$ & $\mathbf{0.573}_{\pm0.00}$ & $\mathbf{0.562}_{\pm0.00}$ \\
\bottomrule
\end{tabular}
\caption{Directional accuracy of MSE\,$+$\,CosDir on ETTh1 by horizon ($H$).}
\label{tab:detailstd_ETTh1}
\end{table*}

\begin{table*}[t]
\centering
\small
\setlength{\tabcolsep}{2pt}
\begin{tabular}{lcccc}
\toprule
Backbone & $H=96$ & $H=192$ & $H=336$ & $H=720$ \\
\midrule
DLinear & $\mathbf{0.603}_{\pm0.01}$ & $\mathbf{0.577}_{\pm0.01}$ & $\mathbf{0.570}_{\pm0.00}$ & $\mathbf{0.563}_{\pm0.00}$ \\
PatchTST & $\mathbf{0.628}_{\pm0.00}$ & $\mathbf{0.624}_{\pm0.00}$ & $\mathbf{0.617}_{\pm0.00}$ & $\mathbf{0.619}_{\pm0.00}$ \\
iTransformer & $\mathbf{0.606}_{\pm0.01}$ & $\mathbf{0.589}_{\pm0.00}$ & $\mathbf{0.583}_{\pm0.00}$ & $\mathbf{0.578}_{\pm0.00}$ \\
TimesNet & $\mathbf{0.650}_{\pm0.00}$ & $\mathbf{0.647}_{\pm0.01}$ & $\mathbf{0.639}_{\pm0.00}$ & $\mathbf{0.624}_{\pm0.02}$ \\
TimeMixer & $\mathbf{0.581}_{\pm0.01}$ & $\mathbf{0.556}_{\pm0.01}$ & $\mathbf{0.551}_{\pm0.01}$ & $\mathbf{0.529}_{\pm0.01}$ \\
Autoformer & $\mathbf{0.603}_{\pm0.02}$ & $\mathbf{0.602}_{\pm0.01}$ & $\mathbf{0.587}_{\pm0.02}$ & $\mathbf{0.587}_{\pm0.01}$ \\
FEDformer & $\mathbf{0.591}_{\pm0.02}$ & $\mathbf{0.565}_{\pm0.01}$ & $\mathbf{0.544}_{\pm0.02}$ & $\mathbf{0.523}_{\pm0.01}$ \\
\bottomrule
\end{tabular}
\caption{Directional accuracy of MSE\,$+$\,CosDir on ETTh2 by horizon ($H$).}
\label{tab:detailstd_ETTh2}
\end{table*}

\begin{table*}[t]
\centering
\small
\setlength{\tabcolsep}{2pt}
\begin{tabular}{lcccc}
\toprule
Backbone & $H=96$ & $H=192$ & $H=336$ & $H=720$ \\
\midrule
DLinear & $0.581_{\pm0.00}$ & $0.579_{\pm0.00}$ & $\mathbf{0.576}_{\pm0.00}$ & $0.569_{\pm0.00}$ \\
PatchTST & $\mathbf{0.570}_{\pm0.00}$ & $\mathbf{0.569}_{\pm0.00}$ & $\mathbf{0.567}_{\pm0.00}$ & $\mathbf{0.562}_{\pm0.00}$ \\
iTransformer & $\mathbf{0.561}_{\pm0.00}$ & $\mathbf{0.556}_{\pm0.00}$ & $\mathbf{0.551}_{\pm0.00}$ & $\mathbf{0.542}_{\pm0.00}$ \\
TimesNet & $\mathbf{0.587}_{\pm0.00}$ & $\mathbf{0.583}_{\pm0.00}$ & $\mathbf{0.578}_{\pm0.00}$ & $\mathbf{0.578}_{\pm0.00}$ \\
TimeMixer & $\mathbf{0.566}_{\pm0.00}$ & $\mathbf{0.562}_{\pm0.00}$ & $\mathbf{0.557}_{\pm0.00}$ & $\mathbf{0.551}_{\pm0.00}$ \\
Autoformer & $\mathbf{0.531}_{\pm0.01}$ & $\mathbf{0.519}_{\pm0.00}$ & $0.523_{\pm0.01}$ & $0.524_{\pm0.01}$ \\
FEDformer & $\mathbf{0.546}_{\pm0.01}$ & $\mathbf{0.545}_{\pm0.01}$ & $0.519_{\pm0.01}$ & $\mathbf{0.513}_{\pm0.01}$ \\
TSMixer & $\mathbf{0.523}_{\pm0.00}$ & $\mathbf{0.521}_{\pm0.00}$ & $\mathbf{0.519}_{\pm0.00}$ & $\mathbf{0.516}_{\pm0.00}$ \\
LightTS & $\mathbf{0.517}_{\pm0.00}$ & $\mathbf{0.517}_{\pm0.00}$ & $\mathbf{0.516}_{\pm0.00}$ & $\mathbf{0.514}_{\pm0.00}$ \\
Pyraformer & $0.517_{\pm0.00}$ & $\mathbf{0.515}_{\pm0.00}$ & $0.513_{\pm0.00}$ & $0.510_{\pm0.00}$ \\
Crossformer & $\mathbf{0.569}_{\pm0.00}$ & $\mathbf{0.568}_{\pm0.00}$ & $\mathbf{0.565}_{\pm0.00}$ & $\mathbf{0.549}_{\pm0.01}$ \\
FiLM & $\mathbf{0.589}_{\pm0.00}$ & $\mathbf{0.587}_{\pm0.00}$ & $\mathbf{0.584}_{\pm0.00}$ & $0.575_{\pm0.00}$ \\
TiDE & $\mathbf{0.532}_{\pm0.00}$ & $\mathbf{0.530}_{\pm0.00}$ & $\mathbf{0.528}_{\pm0.00}$ & $\mathbf{0.525}_{\pm0.00}$ \\
FreTS & $\mathbf{0.547}_{\pm0.00}$ & $\mathbf{0.543}_{\pm0.00}$ & $\mathbf{0.541}_{\pm0.00}$ & $\mathbf{0.537}_{\pm0.00}$ \\
SegRNN & $\mathbf{0.540}_{\pm0.00}$ & $\mathbf{0.542}_{\pm0.00}$ & $\mathbf{0.540}_{\pm0.00}$ & $\mathbf{0.537}_{\pm0.00}$ \\
\bottomrule
\end{tabular}
\caption{Directional accuracy of MSE\,$+$\,CosDir on ETTm1 by horizon ($H$).}
\label{tab:detailstd_ETTm1}
\end{table*}

\begin{table*}[t]
\centering
\small
\setlength{\tabcolsep}{2pt}
\begin{tabular}{lcccc}
\toprule
Backbone & $H=96$ & $H=192$ & $H=336$ & $H=720$ \\
\midrule
DLinear & $0.575_{\pm0.00}$ & $0.573_{\pm0.00}$ & $\mathbf{0.574}_{\pm0.00}$ & $\mathbf{0.574}_{\pm0.00}$ \\
PatchTST & $\mathbf{0.562}_{\pm0.00}$ & $\mathbf{0.556}_{\pm0.00}$ & $\mathbf{0.554}_{\pm0.00}$ & $\mathbf{0.551}_{\pm0.00}$ \\
iTransformer & $\mathbf{0.552}_{\pm0.00}$ & $\mathbf{0.536}_{\pm0.00}$ & $\mathbf{0.529}_{\pm0.00}$ & $\mathbf{0.521}_{\pm0.01}$ \\
TimesNet & $\mathbf{0.603}_{\pm0.00}$ & $\mathbf{0.600}_{\pm0.00}$ & $\mathbf{0.593}_{\pm0.01}$ & $\mathbf{0.596}_{\pm0.01}$ \\
TimeMixer & $\mathbf{0.564}_{\pm0.00}$ & $\mathbf{0.553}_{\pm0.00}$ & $\mathbf{0.550}_{\pm0.01}$ & $\mathbf{0.549}_{\pm0.00}$ \\
Autoformer & $\mathbf{0.545}_{\pm0.01}$ & $\mathbf{0.552}_{\pm0.01}$ & $\mathbf{0.565}_{\pm0.01}$ & $\mathbf{0.568}_{\pm0.02}$ \\
FEDformer & $\mathbf{0.562}_{\pm0.02}$ & $\mathbf{0.556}_{\pm0.02}$ & $0.530_{\pm0.00}$ & $0.525_{\pm0.00}$ \\
TSMixer & $\mathbf{0.511}_{\pm0.00}$ & $\mathbf{0.509}_{\pm0.00}$ & $\mathbf{0.503}_{\pm0.00}$ & $\mathbf{0.501}_{\pm0.00}$ \\
LightTS & $\mathbf{0.509}_{\pm0.00}$ & $\mathbf{0.506}_{\pm0.00}$ & $\mathbf{0.504}_{\pm0.00}$ & $\mathbf{0.503}_{\pm0.00}$ \\
Pyraformer & $0.501_{\pm0.00}$ & $\mathbf{0.501}_{\pm0.00}$ & $\mathbf{0.501}_{\pm0.00}$ & $0.501_{\pm0.00}$ \\
Crossformer & $\mathbf{0.531}_{\pm0.00}$ & $\mathbf{0.511}_{\pm0.01}$ & $\mathbf{0.503}_{\pm0.00}$ & $\mathbf{0.501}_{\pm0.00}$ \\
FiLM & $\mathbf{0.589}_{\pm0.00}$ & $\mathbf{0.587}_{\pm0.00}$ & $\mathbf{0.587}_{\pm0.00}$ & $\mathbf{0.586}_{\pm0.00}$ \\
TiDE & $\mathbf{0.542}_{\pm0.00}$ & $\mathbf{0.539}_{\pm0.00}$ & $\mathbf{0.537}_{\pm0.00}$ & $\mathbf{0.534}_{\pm0.00}$ \\
FreTS & $\mathbf{0.539}_{\pm0.00}$ & $\mathbf{0.529}_{\pm0.00}$ & $\mathbf{0.522}_{\pm0.00}$ & $\mathbf{0.516}_{\pm0.00}$ \\
SegRNN & $0.519_{\pm0.00}$ & $0.517_{\pm0.00}$ & $\mathbf{0.514}_{\pm0.00}$ & $\mathbf{0.511}_{\pm0.00}$ \\
\bottomrule
\end{tabular}
\caption{Directional accuracy of MSE\,$+$\,CosDir on ETTm2 by horizon ($H$).}
\label{tab:detailstd_ETTm2}
\end{table*}

\begin{table*}[t]
\centering
\small
\setlength{\tabcolsep}{2pt}
\begin{tabular}{lcccc}
\toprule
Backbone & $H=96$ & $H=192$ & $H=336$ & $H=720$ \\
\midrule
DLinear & $\mathbf{0.559}_{\pm0.00}$ & $\mathbf{0.564}_{\pm0.00}$ & $\mathbf{0.567}_{\pm0.00}$ & $\mathbf{0.571}_{\pm0.00}$ \\
PatchTST & $\mathbf{0.540}_{\pm0.00}$ & $\mathbf{0.537}_{\pm0.00}$ & $\mathbf{0.536}_{\pm0.00}$ & $\mathbf{0.536}_{\pm0.00}$ \\
iTransformer & $\mathbf{0.551}_{\pm0.01}$ & $\mathbf{0.542}_{\pm0.00}$ & $\mathbf{0.540}_{\pm0.00}$ & $\mathbf{0.535}_{\pm0.00}$ \\
TimesNet & $\mathbf{0.586}_{\pm0.00}$ & $\mathbf{0.582}_{\pm0.00}$ & $\mathbf{0.580}_{\pm0.01}$ & $\mathbf{0.577}_{\pm0.00}$ \\
TimeMixer & $\mathbf{0.554}_{\pm0.00}$ & $\mathbf{0.556}_{\pm0.00}$ & $\mathbf{0.554}_{\pm0.00}$ & $\mathbf{0.557}_{\pm0.00}$ \\
Autoformer & $\mathbf{0.523}_{\pm0.01}$ & $\mathbf{0.534}_{\pm0.02}$ & $\mathbf{0.526}_{\pm0.02}$ & $\mathbf{0.543}_{\pm0.02}$ \\
FEDformer & $\mathbf{0.555}_{\pm0.01}$ & $\mathbf{0.553}_{\pm0.01}$ & $\mathbf{0.550}_{\pm0.02}$ & $\mathbf{0.536}_{\pm0.02}$ \\
\bottomrule
\end{tabular}
\caption{Directional accuracy of MSE\,$+$\,CosDir on Weather by horizon ($H$).}
\label{tab:detailstd_weather}
\end{table*}

\begin{table*}[t]
\centering
\small
\setlength{\tabcolsep}{2pt}
\begin{tabular}{lcccc}
\toprule
Backbone & $H=96$ & $H=192$ & $H=336$ & $H=720$ \\
\midrule
DLinear & $\mathbf{0.750}_{\pm0.00}$ & $\mathbf{0.750}_{\pm0.00}$ & $\mathbf{0.749}_{\pm0.00}$ & $\mathbf{0.753}_{\pm0.00}$ \\
PatchTST & $\mathbf{0.747}_{\pm0.00}$ & $\mathbf{0.740}_{\pm0.00}$ & $\mathbf{0.740}_{\pm0.00}$ & $\mathbf{0.739}_{\pm0.00}$ \\
iTransformer & $\mathbf{0.769}_{\pm0.00}$ & $\mathbf{0.770}_{\pm0.00}$ & $\mathbf{0.773}_{\pm0.00}$ & $\mathbf{0.776}_{\pm0.00}$ \\
TimesNet & $\mathbf{0.764}_{\pm0.00}$ & $\mathbf{0.770}_{\pm0.00}$ & $\mathbf{0.767}_{\pm0.00}$ & $\mathbf{0.775}_{\pm0.00}$ \\
TimeMixer & $\mathbf{0.759}_{\pm0.00}$ & $\mathbf{0.764}_{\pm0.00}$ & $\mathbf{0.766}_{\pm0.00}$ & $\mathbf{0.769}_{\pm0.00}$ \\
Autoformer & $\mathbf{0.572}_{\pm0.05}$ & $\mathbf{0.537}_{\pm0.01}$ & $0.519_{\pm0.02}$ & $\mathbf{0.521}_{\pm0.02}$ \\
FEDformer & $\mathbf{0.756}_{\pm0.01}$ & $\mathbf{0.757}_{\pm0.01}$ & $\mathbf{0.742}_{\pm0.01}$ & $\mathbf{0.704}_{\pm0.02}$ \\
\bottomrule
\end{tabular}
\caption{Directional accuracy of MSE\,$+$\,CosDir on Solar by horizon ($H$).}
\label{tab:detailstd_solar}
\end{table*}

\begin{table*}[t]
\centering
\small
\setlength{\tabcolsep}{2pt}
\begin{tabular}{lcccc}
\toprule
Backbone & $H=96$ & $H=192$ & $H=336$ & $H=720$ \\
\midrule
DLinear & $\mathbf{0.779}_{\pm0.00}$ & $\mathbf{0.783}_{\pm0.00}$ & $\mathbf{0.782}_{\pm0.00}$ & $\mathbf{0.777}_{\pm0.00}$ \\
PatchTST & $0.760_{\pm0.00}$ & $0.764_{\pm0.00}$ & $0.764_{\pm0.00}$ & $0.760_{\pm0.00}$ \\
iTransformer & $\mathbf{0.762}_{\pm0.00}$ & $\mathbf{0.755}_{\pm0.00}$ & $\mathbf{0.756}_{\pm0.00}$ & $\mathbf{0.760}_{\pm0.00}$ \\
TimesNet & $\mathbf{0.817}_{\pm0.01}$ & $\mathbf{0.818}_{\pm0.01}$ & $\mathbf{0.816}_{\pm0.01}$ & $\mathbf{0.806}_{\pm0.02}$ \\
TimeMixer & $\mathbf{0.768}_{\pm0.00}$ & $\mathbf{0.766}_{\pm0.00}$ & $\mathbf{0.765}_{\pm0.00}$ & $\mathbf{0.765}_{\pm0.00}$ \\
Autoformer & $\mathbf{0.752}_{\pm0.02}$ & $\mathbf{0.720}_{\pm0.03}$ & $\mathbf{0.718}_{\pm0.03}$ & $\mathbf{0.716}_{\pm0.03}$ \\
FEDformer & $\mathbf{0.775}_{\pm0.01}$ & $\mathbf{0.768}_{\pm0.01}$ & $\mathbf{0.771}_{\pm0.00}$ & $\mathbf{0.738}_{\pm0.01}$ \\
TSMixer & $\mathbf{0.686}_{\pm0.00}$ & $\mathbf{0.683}_{\pm0.00}$ & $\mathbf{0.679}_{\pm0.00}$ & $\mathbf{0.671}_{\pm0.00}$ \\
LightTS & $\mathbf{0.647}_{\pm0.01}$ & $\mathbf{0.646}_{\pm0.00}$ & $\mathbf{0.643}_{\pm0.00}$ & $\mathbf{0.635}_{\pm0.00}$ \\
Pyraformer & $\mathbf{0.768}_{\pm0.00}$ & $\mathbf{0.773}_{\pm0.00}$ & $\mathbf{0.775}_{\pm0.00}$ & $\mathbf{0.770}_{\pm0.00}$ \\
Crossformer & $\mathbf{0.628}_{\pm0.00}$ & $\mathbf{0.630}_{\pm0.00}$ & $\mathbf{0.632}_{\pm0.00}$ & $\mathbf{0.632}_{\pm0.00}$ \\
FiLM & $\mathbf{0.801}_{\pm0.00}$ & $\mathbf{0.808}_{\pm0.00}$ & $0.796_{\pm0.00}$ & $0.771_{\pm0.00}$ \\
TiDE & $\mathbf{0.737}_{\pm0.00}$ & $\mathbf{0.744}_{\pm0.00}$ & $\mathbf{0.769}_{\pm0.00}$ & -- \\
FreTS & $\mathbf{0.752}_{\pm0.00}$ & $\mathbf{0.760}_{\pm0.00}$ & $\mathbf{0.769}_{\pm0.00}$ & $\mathbf{0.756}_{\pm0.00}$ \\
SegRNN & $\mathbf{0.747}_{\pm0.00}$ & $\mathbf{0.751}_{\pm0.00}$ & $\mathbf{0.749}_{\pm0.00}$ & $\mathbf{0.739}_{\pm0.00}$ \\
\bottomrule
\end{tabular}
\caption{Directional accuracy of MSE\,$+$\,CosDir on Traffic by horizon ($H$).}
\label{tab:detailstd_traffic}
\end{table*}

\begin{table*}[t]
\centering
\small
\setlength{\tabcolsep}{2pt}
\begin{tabular}{lcccc}
\toprule
Backbone & $H=96$ & $H=192$ & $H=336$ & $H=720$ \\
\midrule
DLinear & $\mathbf{0.727}_{\pm0.00}$ & $\mathbf{0.727}_{\pm0.00}$ & $\mathbf{0.726}_{\pm0.00}$ & $\mathbf{0.722}_{\pm0.00}$ \\
PatchTST & $\mathbf{0.745}_{\pm0.00}$ & $\mathbf{0.743}_{\pm0.00}$ & $\mathbf{0.739}_{\pm0.00}$ & $\mathbf{0.731}_{\pm0.00}$ \\
iTransformer & $\mathbf{0.743}_{\pm0.00}$ & $\mathbf{0.734}_{\pm0.00}$ & $\mathbf{0.724}_{\pm0.00}$ & $\mathbf{0.710}_{\pm0.00}$ \\
TimesNet & $\mathbf{0.732}_{\pm0.01}$ & $\mathbf{0.731}_{\pm0.01}$ & $\mathbf{0.730}_{\pm0.01}$ & $\mathbf{0.728}_{\pm0.01}$ \\
TimeMixer & $\mathbf{0.763}_{\pm0.00}$ & $\mathbf{0.754}_{\pm0.00}$ & $\mathbf{0.747}_{\pm0.00}$ & $\mathbf{0.733}_{\pm0.01}$ \\
Autoformer & $\mathbf{0.709}_{\pm0.01}$ & $\mathbf{0.700}_{\pm0.02}$ & $\mathbf{0.698}_{\pm0.02}$ & $\mathbf{0.692}_{\pm0.02}$ \\
FEDformer & $\mathbf{0.713}_{\pm0.02}$ & $\mathbf{0.712}_{\pm0.02}$ & $\mathbf{0.708}_{\pm0.02}$ & $\mathbf{0.689}_{\pm0.01}$ \\
\bottomrule
\end{tabular}
\caption{Directional accuracy of MSE\,$+$\,CosDir on Electricity by horizon ($H$).}
\label{tab:detailstd_electricity}
\end{table*}

\begin{table*}[t]
\centering
\small
\setlength{\tabcolsep}{2pt}
\begin{tabular}{lcccccc}
\toprule
Backbone & $H=12$ & $H=24$ & $H=36$ & $H=48$ & $H=60$ & $H=72$ \\
\midrule
DLinear & $\mathbf{0.549}_{\pm0.00}$ & $\mathbf{0.546}_{\pm0.00}$ & $\mathbf{0.533}_{\pm0.00}$ & $\mathbf{0.523}_{\pm0.00}$ & $\mathbf{0.521}_{\pm0.00}$ & $\mathbf{0.518}_{\pm0.00}$ \\
PatchTST & $\mathbf{0.604}_{\pm0.00}$ & $\mathbf{0.579}_{\pm0.01}$ & $\mathbf{0.555}_{\pm0.00}$ & $\mathbf{0.537}_{\pm0.00}$ & $\mathbf{0.532}_{\pm0.00}$ & $\mathbf{0.526}_{\pm0.00}$ \\
iTransformer & $\mathbf{0.587}_{\pm0.00}$ & $\mathbf{0.556}_{\pm0.01}$ & $\mathbf{0.535}_{\pm0.00}$ & $\mathbf{0.521}_{\pm0.00}$ & $\mathbf{0.516}_{\pm0.00}$ & $\mathbf{0.512}_{\pm0.00}$ \\
TimesNet & $\mathbf{0.518}_{\pm0.00}$ & $\mathbf{0.505}_{\pm0.00}$ & $\mathbf{0.504}_{\pm0.00}$ & $\mathbf{0.503}_{\pm0.00}$ & $\mathbf{0.504}_{\pm0.00}$ & $\mathbf{0.504}_{\pm0.00}$ \\
TimeMixer & $\mathbf{0.588}_{\pm0.01}$ & $\mathbf{0.563}_{\pm0.02}$ & $\mathbf{0.538}_{\pm0.01}$ & $\mathbf{0.524}_{\pm0.00}$ & $\mathbf{0.520}_{\pm0.00}$ & $\mathbf{0.513}_{\pm0.00}$ \\
Autoformer & $\mathbf{0.511}_{\pm0.01}$ & $0.506_{\pm0.00}$ & $\mathbf{0.506}_{\pm0.00}$ & $0.504_{\pm0.00}$ & $\mathbf{0.505}_{\pm0.00}$ & $0.503_{\pm0.00}$ \\
FEDformer & $\mathbf{0.520}_{\pm0.00}$ & $\mathbf{0.513}_{\pm0.00}$ & $\mathbf{0.509}_{\pm0.00}$ & $0.507_{\pm0.00}$ & $0.505_{\pm0.00}$ & $0.504_{\pm0.00}$ \\
\bottomrule
\end{tabular}
\caption{Directional accuracy of MSE\,$+$\,CosDir on RealVar by horizon ($H$).}
\label{tab:detailstd_svar}
\end{table*}

\begin{table*}[t]
\centering
\small
\setlength{\tabcolsep}{2pt}
\begin{tabular}{lcccccc}
\toprule
Backbone & $H=12$ & $H=24$ & $H=36$ & $H=48$ & $H=60$ & $H=72$ \\
\midrule
DLinear & $\mathbf{0.547}_{\pm0.00}$ & $\mathbf{0.543}_{\pm0.00}$ & $\mathbf{0.531}_{\pm0.00}$ & $\mathbf{0.522}_{\pm0.00}$ & $\mathbf{0.520}_{\pm0.00}$ & $\mathbf{0.518}_{\pm0.00}$ \\
PatchTST & $\mathbf{0.590}_{\pm0.00}$ & $\mathbf{0.569}_{\pm0.01}$ & $\mathbf{0.548}_{\pm0.00}$ & $\mathbf{0.532}_{\pm0.00}$ & $\mathbf{0.528}_{\pm0.00}$ & $\mathbf{0.523}_{\pm0.00}$ \\
iTransformer & $\mathbf{0.576}_{\pm0.00}$ & $\mathbf{0.550}_{\pm0.00}$ & $\mathbf{0.532}_{\pm0.00}$ & $\mathbf{0.519}_{\pm0.00}$ & $\mathbf{0.515}_{\pm0.00}$ & $\mathbf{0.512}_{\pm0.00}$ \\
TimesNet & $\mathbf{0.517}_{\pm0.00}$ & $\mathbf{0.506}_{\pm0.00}$ & $\mathbf{0.503}_{\pm0.00}$ & $\mathbf{0.504}_{\pm0.00}$ & $\mathbf{0.504}_{\pm0.00}$ & $\mathbf{0.503}_{\pm0.00}$ \\
TimeMixer & $\mathbf{0.577}_{\pm0.01}$ & $\mathbf{0.558}_{\pm0.01}$ & $\mathbf{0.534}_{\pm0.00}$ & $\mathbf{0.522}_{\pm0.00}$ & $\mathbf{0.518}_{\pm0.00}$ & $\mathbf{0.512}_{\pm0.00}$ \\
Autoformer & $\mathbf{0.513}_{\pm0.00}$ & $0.507_{\pm0.00}$ & $\mathbf{0.506}_{\pm0.00}$ & $0.504_{\pm0.00}$ & $\mathbf{0.504}_{\pm0.00}$ & $\mathbf{0.504}_{\pm0.00}$ \\
FEDformer & $\mathbf{0.520}_{\pm0.00}$ & $\mathbf{0.512}_{\pm0.00}$ & $\mathbf{0.509}_{\pm0.00}$ & $\mathbf{0.508}_{\pm0.00}$ & $\mathbf{0.505}_{\pm0.00}$ & $\mathbf{0.504}_{\pm0.00}$ \\
\bottomrule
\end{tabular}
\caption{Directional accuracy of MSE\,$+$\,CosDir on RealVol by horizon ($H$).}
\label{tab:detailstd_svol20log}
\end{table*}

\begin{table*}[t]
\centering
\small
\setlength{\tabcolsep}{2pt}
\begin{tabular}{lcccccc}
\toprule
Backbone & $H=12$ & $H=24$ & $H=36$ & $H=48$ & $H=60$ & $H=72$ \\
\midrule
DLinear & $\mathbf{0.532}_{\pm0.01}$ & $\mathbf{0.538}_{\pm0.00}$ & $\mathbf{0.538}_{\pm0.01}$ & $\mathbf{0.542}_{\pm0.00}$ & $\mathbf{0.538}_{\pm0.01}$ & $\mathbf{0.531}_{\pm0.01}$ \\
PatchTST & $\mathbf{0.572}_{\pm0.00}$ & $\mathbf{0.567}_{\pm0.01}$ & $\mathbf{0.571}_{\pm0.00}$ & $\mathbf{0.563}_{\pm0.01}$ & $\mathbf{0.568}_{\pm0.00}$ & $\mathbf{0.555}_{\pm0.00}$ \\
iTransformer & $\mathbf{0.566}_{\pm0.00}$ & $\mathbf{0.549}_{\pm0.00}$ & $\mathbf{0.542}_{\pm0.00}$ & $\mathbf{0.531}_{\pm0.00}$ & $\mathbf{0.526}_{\pm0.00}$ & $\mathbf{0.519}_{\pm0.00}$ \\
TimesNet & $\mathbf{0.510}_{\pm0.01}$ & $\mathbf{0.506}_{\pm0.00}$ & $\mathbf{0.501}_{\pm0.00}$ & $\mathbf{0.501}_{\pm0.00}$ & $\mathbf{0.500}_{\pm0.00}$ & $\mathbf{0.500}_{\pm0.00}$ \\
TimeMixer & $\mathbf{0.573}_{\pm0.01}$ & $\mathbf{0.553}_{\pm0.01}$ & $\mathbf{0.552}_{\pm0.01}$ & $\mathbf{0.537}_{\pm0.01}$ & $\mathbf{0.530}_{\pm0.01}$ & $\mathbf{0.518}_{\pm0.00}$ \\
Autoformer & $\mathbf{0.503}_{\pm0.00}$ & $\mathbf{0.502}_{\pm0.00}$ & $\mathbf{0.503}_{\pm0.00}$ & $\mathbf{0.500}_{\pm0.00}$ & $0.500_{\pm0.00}$ & $\mathbf{0.502}_{\pm0.00}$ \\
FEDformer & $\mathbf{0.506}_{\pm0.00}$ & $\mathbf{0.506}_{\pm0.00}$ & $\mathbf{0.506}_{\pm0.00}$ & $\mathbf{0.506}_{\pm0.00}$ & $\mathbf{0.505}_{\pm0.00}$ & $\mathbf{0.503}_{\pm0.00}$ \\
\bottomrule
\end{tabular}
\caption{Directional accuracy of MSE\,$+$\,CosDir on RealVol60 by horizon ($H$).}
\label{tab:detailstd_svol60}
\end{table*}

\begin{table*}[t]
\centering
\small
\setlength{\tabcolsep}{2pt}
\begin{tabular}{lcccccc}
\toprule
Backbone & $H=12$ & $H=24$ & $H=36$ & $H=48$ & $H=60$ & $H=72$ \\
\midrule
DLinear & $\mathbf{0.538}_{\pm0.00}$ & $\mathbf{0.536}_{\pm0.00}$ & $\mathbf{0.524}_{\pm0.00}$ & $\mathbf{0.518}_{\pm0.00}$ & $\mathbf{0.514}_{\pm0.00}$ & $\mathbf{0.513}_{\pm0.00}$ \\
PatchTST & $\mathbf{0.574}_{\pm0.00}$ & $\mathbf{0.556}_{\pm0.01}$ & $\mathbf{0.537}_{\pm0.00}$ & $\mathbf{0.525}_{\pm0.00}$ & $\mathbf{0.521}_{\pm0.00}$ & $\mathbf{0.518}_{\pm0.00}$ \\
iTransformer & $\mathbf{0.566}_{\pm0.00}$ & $\mathbf{0.541}_{\pm0.00}$ & $\mathbf{0.526}_{\pm0.00}$ & $\mathbf{0.515}_{\pm0.00}$ & $\mathbf{0.512}_{\pm0.00}$ & $\mathbf{0.509}_{\pm0.00}$ \\
TimesNet & $\mathbf{0.517}_{\pm0.00}$ & $\mathbf{0.511}_{\pm0.00}$ & $0.503_{\pm0.00}$ & $\mathbf{0.505}_{\pm0.00}$ & $0.502_{\pm0.00}$ & $\mathbf{0.503}_{\pm0.00}$ \\
TimeMixer & $\mathbf{0.559}_{\pm0.01}$ & $\mathbf{0.546}_{\pm0.01}$ & $\mathbf{0.524}_{\pm0.00}$ & $\mathbf{0.514}_{\pm0.00}$ & $\mathbf{0.511}_{\pm0.00}$ & $\mathbf{0.507}_{\pm0.00}$ \\
Autoformer & $\mathbf{0.511}_{\pm0.00}$ & $0.507_{\pm0.00}$ & $0.506_{\pm0.00}$ & $\mathbf{0.506}_{\pm0.00}$ & $0.505_{\pm0.00}$ & $0.505_{\pm0.00}$ \\
FEDformer & $\mathbf{0.517}_{\pm0.00}$ & $\mathbf{0.510}_{\pm0.00}$ & $\mathbf{0.508}_{\pm0.00}$ & $\mathbf{0.508}_{\pm0.00}$ & $\mathbf{0.506}_{\pm0.00}$ & $\mathbf{0.506}_{\pm0.00}$ \\
\bottomrule
\end{tabular}
\caption{Directional accuracy of MSE\,$+$\,CosDir on StockVol by horizon ($H$).}
\label{tab:detailstd_stock_vol}
\end{table*}

\begin{table*}[t]
\centering
\small
\setlength{\tabcolsep}{2pt}
\begin{tabular}{lcccccc}
\toprule
Backbone & $H=12$ & $H=24$ & $H=36$ & $H=48$ & $H=60$ & $H=72$ \\
\midrule
DLinear & $\mathbf{0.521}_{\pm0.00}$ & $\mathbf{0.512}_{\pm0.00}$ & $\mathbf{0.509}_{\pm0.00}$ & $\mathbf{0.507}_{\pm0.00}$ & $\mathbf{0.507}_{\pm0.00}$ & $0.505_{\pm0.00}$ \\
PatchTST & $\mathbf{0.520}_{\pm0.00}$ & $\mathbf{0.510}_{\pm0.00}$ & $\mathbf{0.507}_{\pm0.00}$ & $\mathbf{0.506}_{\pm0.00}$ & $\mathbf{0.506}_{\pm0.00}$ & $\mathbf{0.505}_{\pm0.00}$ \\
iTransformer & $\mathbf{0.519}_{\pm0.00}$ & $\mathbf{0.509}_{\pm0.00}$ & $\mathbf{0.506}_{\pm0.00}$ & $\mathbf{0.505}_{\pm0.00}$ & $\mathbf{0.504}_{\pm0.00}$ & $0.503_{\pm0.00}$ \\
TimesNet & $\mathbf{0.520}_{\pm0.00}$ & $\mathbf{0.511}_{\pm0.00}$ & $\mathbf{0.507}_{\pm0.00}$ & $\mathbf{0.505}_{\pm0.00}$ & $0.504_{\pm0.00}$ & $0.503_{\pm0.00}$ \\
TimeMixer & $\mathbf{0.521}_{\pm0.00}$ & $\mathbf{0.511}_{\pm0.00}$ & $0.507_{\pm0.00}$ & $\mathbf{0.505}_{\pm0.00}$ & $0.504_{\pm0.00}$ & $0.503_{\pm0.00}$ \\
Autoformer & $0.520_{\pm0.00}$ & $0.510_{\pm0.01}$ & $\mathbf{0.509}_{\pm0.00}$ & $0.504_{\pm0.00}$ & $\mathbf{0.507}_{\pm0.00}$ & $\mathbf{0.507}_{\pm0.00}$ \\
FEDformer & $\mathbf{0.522}_{\pm0.00}$ & $\mathbf{0.512}_{\pm0.00}$ & $\mathbf{0.509}_{\pm0.00}$ & $0.507_{\pm0.00}$ & $0.506_{\pm0.00}$ & $0.505_{\pm0.00}$ \\
\bottomrule
\end{tabular}
\caption{Directional accuracy of MSE\,$+$\,CosDir on AbsRet by horizon ($H$).}
\label{tab:detailstd_sabsret}
\end{table*}

\end{document}